\documentclass[lettersize,journal]{IEEEtran}
\usepackage{amsmath,amsfonts,bm}
\usepackage{amssymb}
\usepackage{algorithmic}
\usepackage{algorithm}
\usepackage{array}
\usepackage[caption=false,font=normalsize,labelfont=sf,textfont=sf]{subfig}
\usepackage{textcomp}
\usepackage{stfloats}
\usepackage{url}
\usepackage{hyperref}
\usepackage{verbatim}
\usepackage{graphicx}
\usepackage{cite}
\usepackage{mathtools}
\usepackage{etoolbox}
\DeclarePairedDelimiter\floor{\lfloor}{\rfloor}
\newtheorem{theorem}{Theorem}[section]

\newtheorem{lemma}[theorem]{Lemma}

\newtheorem{assumption}[theorem]{Assumption}

\begin{document}

\title{QC-ODKLA: Quantized and Communication-Censored Online Decentralized Kernel Learning via Linearized ADMM}

\author{Ping Xu, Yue Wang,~\IEEEmembership{Senior Member,~IEEE,} Xiang Chen, Zhi Tian,~\IEEEmembership{Fellow,~IEEE}
        % <-this % stops a space
\thanks{This work was partly supported by the National Science Foundation of the US (Grant \#1741338, \#1939553, \#2003211, \#2128596, \#2136202), and the Virginia Research Investment Fund (Commonwealth Cyber Initiative Grant \#223996).}% <-this % stops a space
%\thanks{Manuscript received April 19, 2021; revised August 16, 2021.}
}

%\markboth{Journal of \LaTeX\ Class Files,~Vol.~ , No.~8, Ju~2022}%

%\IEEEpubid{0000--0000/00\$00.00~\copyright~2022 IEEE}

\maketitle

\begin{abstract}
This paper focuses on online kernel learning over a decentralized network. Each agent in the network receives continuous streaming data locally and works collaboratively to learn a nonlinear prediction function that is globally optimal in the reproducing kernel Hilbert space with respect to the total instantaneous costs of all agents. In order to circumvent the curse of dimensionality issue in traditional online kernel learning, we utilize random feature (RF) mapping to convert the non-parametric kernel learning problem into a fixed-length parametric one in the RF space. We then propose a novel learning framework named Online Decentralized Kernel learning via Linearized ADMM (ODKLA) to efficiently solve the online decentralized kernel learning problem. To further improve the communication efficiency, we add the quantization and censoring strategies in the communication stage and develop the Quantized and Communication-censored ODKLA (QC-ODKLA) algorithm. We theoretically prove that both ODKLA and QC-ODKLA can achieve the optimal sublinear regret $\mathcal{O}(\sqrt{T})$ over $T$ time slots. Through numerical experiments, we evaluate the learning effectiveness, communication, and computation efficiencies of the proposed methods.  
\end{abstract}

\begin{IEEEkeywords}
Decentralized online kernel learning, random feature mapping, linearized ADMM, communication-censoring, quantization.
\end{IEEEkeywords}

\section{Introduction}
\label{sec:intro}

Decentralized online learning has been widely studied in the last decades, mostly motivated by its broad applications in networked multi-agent systems, such as wireless sensor networks, robotics, and internet of things, etc~\cite{liang2012distributed, ren2007information}. In these systems, a number of agents collect their own online streaming data and aim to learn a common functional model through local information exchange. %The term \textit{local information exchange} indicates that each agent in the network only communicates with its neighbors within the one-hop local communication range to save transmission power.
%At each time, agents learn the functional model based on historical information at hand and currently received data samples, without known future information on cost functions. 
This objective is usually achieved by decentralized online convex optimization~\cite{mateos2014distributed, xu2015online, koppel2015saddle, zhao2019decentralized, sharma2020distributed}. With an online gradient descent based algorithm~\cite{zinkevich2003online}, or through online alternating direction method of multipliers (ADMM)~\cite{xu2015online}, a static regret $\mathcal{O}(\sqrt{T})$ can be achieved over a time horizon $T$. Further, if the cost functions are strictly convex, an efficient algorithm based on the Newton method achieves a regret bound of $\mathcal{O}(\log T)$~\cite{hazan2007logarithmic}. In addition to static environments, online learning in dynamic environments has attracted more and more attentions recently~\cite{kamp2014communication, shahrampour2017distributed, asghari2020regret, dixit2020online, zhao2021proximal}. However, all these works assume that the functional model to be learned by agents is linear, which may not be always true in practical applications. 

Motivated by the universality of kernel methods in approximating nonlinear functions, this paper aims to solve the decentralized online kernel learning problem where the common function to be learned by agents is assumed to be nonlinear and belong to the reproducing kernel Hilbert space (RKHS). However, directly applying kernel methods for decentralized online learning is formidably challenging because they adopt nonparametric models where the number of model variables grows proportionally to the data size, which incurs the curse of dimensionality issue when data size goes large as time evolves. In addition, the data-dependent decision variables prevent consensus optimization when the data sizes vary at different agents and across time as well as under certain circumstances where raw data exchange is prohibited~\cite{xu2021coke}. 

To alleviate the computational complexity of kernel methods, various dimensionality reduction techniques have been developed, including stochastic approximation~\cite{gu2018asynchronous}, restricting the number of function parameters~\cite{le2016nonparametric, koppel2017parsimonious}, and approximating the kernel during training~\cite{richard2008online, dai2014scalable, rahimi2008random, nguyen2017large}. Among them, random feature (RF) mapping methods~\cite{rahimi2008random, dai2014scalable, nguyen2017large} not only circumvent the curse of dimensionality problem but also enable consensus optimization without any raw data exchange among agents, which makes them popular in many decentralized kernel learning works, including batch-form learning~\cite{xu2021coke, richards2020decentralised} and online streaming learning~\cite{bouboulis2018online, shen2021distributed, hong2021distributed}.    

%thanks to their ability to transform the original nonparametric data-dependent learning problem into a parametric fixed small size data-independent learning problem by approximating the kernel with a fixed (small) number of random features~\cite{rahimi2008random, dai2014scalable, nguyen2017large}. Note that RF mapping not only circumvents the curse of dimensionality problem but also enables consensus optimization and thus is adopted in many decentralized kernel learning works, including batch-form learning~\cite{xu2021coke, richards2020decentralised} and online streaming learning~\cite{bouboulis2018online, shen2021distributed, hong2021distributed}.    
 
Another key problem in decentralized learning is that it relies on iterative local communications for computational feasibility and efficiency. This incurs frequent communications among agents to exchange their locally computed updates of the shared learning model, which can cause tremendous communication overhead in terms of both link bandwidth and transmission power. Therefore, communication-efficient algorithms are desired in decentralized learning. To improve the communication efficiency, we can harness the function smoothness or the Nesterov gradient to achieve fast convergence~\cite{qu2017harnessing, qu2019accelerated}, transmit the compressed information by quantization~\cite{zhu2016quantized, zhang2019quantized, shen2021distributed} or sparsification~\cite{NIPS2018_7697, harrane2018reducing}, randomly select a number of nodes for broadcasting/communication, and operate asynchronous updating to reduce the number of transmissions per iteration~\cite{li2014communication, arablouei2015analysis, mcmahan2016communication, yin2018communication, NIPS2019_8694}. In contrast to random node selection, a more intuitive way is to evaluate the importance of a message in order to avoid unnecessary transmissions. This is usually implemented by adopting a communication censoring/event-triggering scheme to adaptively decide if a message is informative enough to be transmitted during the iterative optimization process~\cite{chen2018lag, liu2019communication, li2019cola, xu2021coke, cao2020decentralized}.

In this article, we thus focus on the decentralized online kernel learning problem in networked multi-agent systems and aim to develop both communication- and computation-efficient algorithms. We first utilize RF mapping to transform the original nonparametric data-dependent learning problem into a parametric fixed-size data-independent learning problem to circumvent the curse of dimensionality issue in traditional kernel methods and enable consensus optimization in a decentralized setting in the RF space. Different from existing gradient descent based method~\cite{bouboulis2018online, shen2021distributed} or standard ADMM algorithm~\cite{hong2021distributed}, we propose to solve the decentralized kernel learning problem by linearized ADMM and develop the \textbf{O}nline \textbf{D}ecentralized \textbf{K}ernel learning via \textbf{L}inearized \textbf{A}DMM (ODKLA) algorithm. In ODKLA, the local cost function of each agent is replaced by its first-order approximation centered at the current iterate and results in a closed-form primal update if the local cost function is convex. In this way, the computation efficiency of ODKLA is improved compared with standard ADMM where the primal update requires to solve a suboptimization problem every time while still enjoying fast convergence speed. To further reduce the communication cost, we develop the \textbf{Q}uantized and \textbf{C}ommunication-censored \textbf{O}nline \textbf{D}ecentralized \textbf{K}ernel learning via \textbf{L}inearized \textbf{A}DMM (QC-ODKLA) algorithm by introducing a communication censoring strategy and a quantization strategy. The communication censoring strategy allows each agent to autonomously skip unnecessary communications when its local update is not informative enough for transmission, while the quantization strategy restricts the total number of bits transmitted in the learning process. The communication efficiency can be boosted at almost no sacrifice to the learning performance. %Quantization is also considered in~\cite{shen2021distributed} with the aim of improving communication efficiency. 
%While both~\cite{shen2021distributed} and \cite{hong2021distributed} study multi-kernel learning, which shows improved learning performance over single kernel-based algorithms, the algorithms we proposed in this article can also be easily extended to multi-kernel learning. 
Our key contributions are summarized as follows.
\begin{itemize}
	\item We develop the ODKLA that utilizes linearized ADMM to solve the online decentralized multi-agent kernel learning problem in the RF space. ODKLA is fully decentralized and does not involve solving sub-optimization problems, which is thus more computationally efficient than standard ADMM. Moreover, ODKLA is essentially a variant of the higher-order ADMM and thus achieves faster convergence compared with the diffusion-based first-order gradient descent methods~\cite{bouboulis2018online}.
	
	\item Utilizing both communication-censoring and quantization strategies, we develop the QC-ODKLA algorithm, which achieves desired learning performance given limited communication resources and energy supply. When both strategies are absent, QC-ODKLA degenerates to ODKLA.  
	
	\item In addition, we analyze the regret bound of QC-ODKLA. We show that when all techniques are adopted (linearized ADMM, quantization, and communication censoring), QC-ODKLA is still able to achieve the optimal sublinear regret $\mathcal{O}(\sqrt{T})$ over $T$ time slots under mild conditions, i.e., the communication censoring thresholds should be decaying.

	\item Finally, we test the performance of our proposed ODKLA and QC-ODKLA algorithms on extensive real datasets. The results corroborate that both ODKLA and QC-ODKLA exhibit attractive learning performance and computation  efficiency, while QC-ODKLA is highly communication-efficient. Such salient features make it an attractive solution for broad applications where decentralized learning from streaming data is at its core. 
\end{itemize}

The remaining of this paper is organized as follows. Section \ref{sec:prelim} provides some preliminaries for decentralized kernel learning. Section \ref{sec:prostate} formulates the online decentralized kernel learning problem. Section \ref{sec:algdev} develops the online decentralized kernel learning algorithms, including both ODKLA and QC-ODKLA. Section \ref{sec:regret} presents the theoretical results. Section \ref{sec:exper} tests the proposed methods by real datasets. Concluding remarks are summarized in Section \ref{sec:con}.

\textbf{Notation.} $\mathbb{R}$ denotes the set of real numbers. $\|\cdot\|_2$ denotes the Euclidean norm of vectors and $\|\cdot\|_F$ denotes the Frobenius norm of matrices. $|\cdot|$ denotes the cardinality of a set. $\mathbf{A}$ denotes a matrix, $\mathbf{a}$ denotes a vector, and $a$ denotes a scalar.

\section{Preliminaries}
\label{sec:prelim}
\subsection{Network and communication models}
\label{subsec:net_com}
\textbf{Network Model.} Consider a bidirectionally connected network of $N$ agents and $r$ arcs, whose underlying undirected communication graph is denoted as $\mathcal{G}=(\mathcal{N}, \mathcal{A}) $, where $\mathcal{N}$ is the set of agents with cardinality $|\mathcal{N}|= N$ and $\mathcal{A}$ is the set of undirected arcs with cardinality $|\mathcal{A}|=r$. Two agents $i$ and $j$ are called as neighbors when $(i,j)\in \mathcal{A}$ and, by the symmetry of the network, $(j,i)\in \mathcal{A}$. For agent $i$, its one-hop neighbors are in the set $\mathcal{N}_i=\{j|(j,i)\in \mathcal{A}\}$ with cardinality $|\mathcal{N}_i|$, which is also known as the degree $d_i$ of agent $i$. The degree matrix of the communication graph is $\bm{D}\in\mathbb{R}^{N\times N}$ which is diagonal with the $i$th diagonal element being $d_i, \forall i$. Define the symmetric adjacency matrix associated with the communication graph as $\bm{W}\in\mathbb{R}^{N\times N}$, whose $(i,j)$th entry is 1 if agent $i$ and $j$ are neighbors or 0 otherwise. Define the unsigned incidence matrix and the signed incidence matrix of the communication graph as $\mathbf{S}_{+}\in\mathbb{R}^{N\times 2r}$ and $\mathbf{S}_{-}\in\mathbb{R}^{N\times 2r}$, respectively. According to~\cite{chung1997spectral}, we have 
\begin{equation}
	\begin{split}
		\textstyle 	\bm{D} + \bm{W} &= \frac{1}{2} \mathbf{S}_{+}\mathbf{S}_{+}^\top,\\
		\textstyle 	\bm{D} - \bm{W} &= \frac{1}{2} \mathbf{S}_{-}\mathbf{S}_{-}^\top. \nonumber
	\end{split}
\end{equation}
 
\textbf{Communication Model.} In this paper, we consider synchronous communications. That is, the iterative process of algorithm implementation consists of three stages: communication, observation, and computation. In the communication stage, each agent broadcasts its state variable to its neighbors and receives state variables from its neighbors according to the communication censoring rule, which shall be introduced later. After communicating with its neighbors, each agent collects its streaming data and formulates its own local objective function in the observation stage. In the computation stage, each agent carries out local updates based on the observed data, local objective function, and state variables.  

\subsection{Random feature mapping}
\label{subsec:rfmapping}
Random feature (RF) mapping is proposed to make kernel methods scalable for large datasets~\cite{rahimi2008random}.  For a shift-invariant kernel that satisfies $\kappa(\mathbf{x}_{t},\mathbf{x}_\tau) = \kappa(\mathbf{x}_{t}-\mathbf{x}_\tau),\;\forall t,\;\forall\tau$, if $\kappa(\mathbf{x}_{t}-\mathbf{x}_\tau)$ is absolutely integrable, then its Fourier transform $p_\kappa(\bm{\omega})$ is guaranteed to be nonnegative ($p_\kappa(\bm{\omega})\geq 0$), and hence can be viewed as its probability density function (pdf) when $\kappa$ is scaled to satisfy $\kappa(0)=1$~\cite{bochner2005harmonic}. Therefore, we have  
\begin{equation}
	\label{eq:kern_fouri}
	\begin{split}
		\textstyle  \kappa(\mathbf{x}_{t},\mathbf{x}_\tau) 	&= \int  p_\kappa(\bm{\omega})e^{j\bm{\omega}^\top(\mathbf{x}_{t}-\mathbf{x}_\tau)}d\bm{\omega} \\
		& = \mathbb{E}_{\bm{\omega}}[\phi(\mathbf{x}_{t},\bm{\omega})\phi^\ast(\mathbf{x}_{\tau},\bm{\omega})],
	\end{split}
\end{equation} 
where $\mathbb{E}$ denotes the expectation operator, $\phi(\mathbf{x},\bm{\omega}):= e^{j\bm{\omega}^\top \mathbf{x}}$ with $\bm{\omega}\in\mathbb{R}^d$, and $\ast$ is the complex conjugate operator. In \eqref{eq:kern_fouri}, the first equality is the result of the Fourier inversion theorem, and the second equality arises by viewing $p_{\kappa}(\bm{\omega})$ as the pdf of $\bm{\omega}$. In this paper, we adopt a Gaussian kernel $\kappa(\mathbf{x}_{t},\mathbf{x}_\tau) =\rm{exp}(-\|\mathbf{x}_{t}-\mathbf{x}_{\tau}\|_2^2/(2\sigma^2))$, whose pdf is a normal distribution with $p_\kappa(\bm{\omega})\sim\mathbf{N}(\bf{0},\sigma^{-2}\bf{I})$.  
The main idea of the RF mapping method is to approximate the kernel function $\kappa(\mathbf{x}_{t},\mathbf{x}_\tau)$ by the sample average
\begin{equation}
	\label{eq:kernel_map_L}
	\begin{split}
		\textstyle \hat{\kappa}_L(\mathbf{x}_{t},\mathbf{x}_\tau)   &:=\frac{1}{L}\sum_{l=1}^L \phi(\mathbf{x}_t,\bm{\omega}_l)\phi^\ast(\mathbf{x}_\tau,\bm{\omega}_l),\\
		%&:=\bm{\phi}_L^\dagger (\mathbf{\mathbf{x}}_{\tau})\bm{\phi}_L(\mathbf{\mathbf{x}}_{t}),
	\end{split}
\end{equation}
where $\{\bm{\omega}_l\}_{l=1}^L$ are randomly drawn from the distribution $p_\kappa(\bm{\omega})$, and $^\ast$ is the conjugate operator. For implementation, the following real-valued mapping is usually adopted:
\begin{equation}
	\label{eq:realRF}
	\phi(\mathbf{x},\bm{\omega})=[\cos(\bm{\omega} ^\top\mathbf{x}),\sin(\bm{\omega}^\top\mathbf{x})]^\top.
\end{equation}

\section{Problem Statement} 
\label{sec:prostate}
Consider the network model described in Section \ref{subsec:net_com}, each agent in the network only has access to its locally observed data composed of independently and identically distributed (i.i.d) input-label pairs $\{\mathbf{x}_{i,t},y_{i,t}\}_{t=1}^{T}$ obeying an unknown probability distribution $p$ on $\mathcal{X}\times \mathcal{Y}$, with $\mathbf{x}_{i,t} \in \mathbb{R}^d$ and $y_{i,t} \in \mathbb{R}$. The decentralized learning task is to find a nonlinear prediction function $f$ such that $y_{i,t}=f(\mathbf{x}_{i,t})+e_{i,t}$ for $\{\{\mathbf{x}_{i,t},y_{i,t}\}_{t=1}^{T}\}_{i=1}^N$, where the error term $e_{i,t}$ is minimized accordingly to certain optimality metric. This is usually achieved by minimizing the empirical risk:
\begin{equation}
	\label{eq:empirical_risk}
	\textstyle	f^\star = \underset{f \in \Omega }{\text{arg min}} \quad  \sum_{i=1}^N \sum_{t=1}^{T} \ell(f(\mathbf{x}_{i,t}),y_{i,t}) + \lambda \|f\|_{\Omega}^2,
\end{equation}
where $\ell(\cdot, \cdot)$ is a nonnegative loss function, $\Omega$ is the function space $f$ belongs to, and $\lambda>0$ is a regularization parameter that controls over-fitting. For regression problems, a common loss function is the quadratic loss. For binary classifications, the common loss functions are the hinge loss $\ell(y,\hat{y})=\max(0,1-y\hat{y})$ and the logistic loss $\ell(y,\hat{y})=\log(1+e^{-y\hat{y}})$.

Assume $f$ belongs to the RKHS $\mathcal{H}:=\{f|f(\mathbf{x})=\sum_{t=1}^\infty \alpha_t \kappa(\mathbf{x},\mathbf{x}_t)\}$ induced by a shift-invariant positive semidefinite kernel $\kappa(\mathbf{x},\mathbf{x}_t):\mathbb{R}^d\times\mathbb{R}^d\rightarrow \mathbb{R}$, and adopt the RF mapping method described in Section \ref{subsec:rfmapping}. Then, the function $f^\star$ to be learned in \eqref{eq:empirical_risk} can be approximated by the following representation:
\begin{equation}
	\label{eq:minimizer_RF}
	\textstyle  \hat{f}^\star(\mathbf{x})=\bm{\theta}^\top\bm{\phi}_L(\mathbf{x}),
\end{equation}
where $\bm{\theta}\in\mathbb{R}^{2L}$ is the decision vector to be learned in the RF space, and $\bm{\phi}_L(\mathbf{x})$ is the mapped data in the RF space using \eqref{eq:realRF}:
\begin{equation}
	\label{eq:RF_map}
	\textstyle  \bm{\phi}_L(\mathbf{x}):=\sqrt{\frac{1}{L}}[\phi(\mathbf{x},\bm{\omega}_1),\dots, \phi(\mathbf{x},\bm{\omega}_L)]^\top.
\end{equation}

With the approximation \eqref{eq:minimizer_RF}, the decentralized kernel learning problem is formulated as
\begin{equation}
	\label{eq:decentr_theta}
	\begin{split}
		&\underset{\{\bm{\theta}_i, \bm{z}_{ij} \}}{\text{min}} \quad  \sum_{i=1}^N \Big[ \sum_{t=1}^{T}  \ell(\bm{\theta}_i^\top\bm{\phi}_L(\mathbf{x}_{i,t}),y_{i,t})+\frac{\lambda}{N}\|\bm{\theta}_i\|^2 \Big]\\
		&\quad \text{s.t.}  \quad \quad  \bm{\theta}_i=\bm{z}_{ij},\; \bm{\theta}_j=\bm{z}_{ij}, \qquad \forall (i,j)\in \mathcal{A},
	\end{split}
\end{equation} 
where $\bm{\theta}_i$ is the local copy of the global parameter $\bm{\theta}$ associated with each agent $i$. The constraint in \eqref{eq:decentr_theta} enforces the consensus constraint on neighboring agents $i$ and $j$ using an auxiliary variable $\bm{z}_{ij}$. The optimization problem can then be solved using DKLA proposed in \cite{xu2021coke}. A communication-censored algorithm (COKE) is also proposed in \cite{xu2021coke} to improve the communication efficiency of DKLA. 

However, both DKLA and COKE operate in batch form when all data are available. Whereas in many real-life applications, function learning tasks are expected to perform in an online fashion with sequentially arriving data. In this article, we consider the case that each agent collects the data points $\{\mathbf{x}_{i,t},y_{i,t}\}_{t=1}^{T}, \forall i$ in an online fashion, and the parameter is estimated based on instantaneous data samples. To achieve an optimal sublinear regrets from the optimal performance of \eqref{eq:decentr_theta}, we customize the general online decentralized alternating direction method of multipliers algorithm proposed in \cite{xu2015online} to decentralized online kernel learning to efficiently solve the online kernel learning problem over a decentralized network. At every time $t$, decentralized online kernel learning (approximately) solves an optimization problem to obtain the update $\bm{\theta}_{i,t+1}$ from the current decision $\bm{\theta}_{i,t}$ and the newly arrived data: 
%
%Note: I think $\theta_{i,t}$ has not been updated. Hence, some description can be added here to serve as the definition as well :
\begin{equation}
	\label{eq:online_opt}
	\begin{split}
		\textstyle   &\underset{\{\bm{\theta}_i,\bm{z}_{ij}\}}{\text{argmin} } \; \sum_{i=1}^N \mathcal{L}_{i,t} (\bm{\theta}_i)  + \frac{\eta_{t}}{2}\sum_{i=1}^N \|\bm{\theta}_i - \bm{\theta}_{i,t}\|^2 \\
		&\text{s.t.} \quad  \bm{\theta}_i = \bm{z}_{ij}, \bm{\theta}_j = \bm{z}_{ij}, \; \forall (i,j) \in \mathcal{A},
	\end{split}
\end{equation}
where $\mathcal{L}_{i,t} (\bm{\theta}_i):= \ell(\bm{\theta}_i^\top\bm{\phi}_L(\mathbf{x}_{i,t}),y_{i,t})+\frac{\lambda}{N}\|\bm{\theta}_i\|^2$ is the local instantaneous cost function dependent of the new data only, whereas $\bm{\theta}_{i,t}$ captures the influence of all the past data. 

In the next section, we first propose a computation-efficient algorithm to solve \eqref{eq:online_opt}. We then utilize communication-censoring and quantization strategies to improve the communication efficiency of the proposed algorithm. 

\section{Algorithm Development}
\label{sec:algdev}
In this section, we first utilize linearized ADMM to efficiently solve \eqref{eq:online_opt} and then add the censoring and quantization techniques to develop a communication-efficient decentralized online kernel learning algorithm.

For notational clarity, we define $\bm{\Theta} = [\bm{\theta}_1^\top;\bm{\theta}_2^\top;\dots;\bm{\theta}_N^\top]\in\mathbb{R}^{N\times 2L}$ that contains all the local copies $\bm{\theta}_i$ and  $\bm{Z} = [\cdots;\bm{z}_{ij}^\top;\cdots]\in\mathbb{R}^{2r\times 2L}$. We further define the aggregated function as $\mathcal{L}_t (\bm{\Theta}):=\sum_{i=1}^N \mathcal{L}_{i,t} (\bm{\theta}_i)$. With these definitions, we rewrite \eqref{eq:online_opt} in a matrix form for the $\bm{\Theta}_{t+1}$ update:
\begin{equation}
	\label{eq:online_opt_matrix}
	\begin{split}
		& 	\textstyle  \underset{\{\bm{\Theta},\bm{Z}\}}{\text{argmin} } \; \mathcal{L}_{t} (\bm{\Theta})+ \frac{\eta_{t}}{2} \|\bm{\Theta} - \bm{\Theta}_{t}\|^2 \\
		&	\textstyle \qquad \text{s.t.} \qquad   \bm{A}\bm{\Theta} + \bm{B}\bm{Z} = \mathbf{0}_{4r\times 2L},
	\end{split}
\end{equation}
where $\bm{A} = \frac{1}{2}[ \mathbf{S}_{+}^\top +\mathbf{S}_{-}^\top ; \mathbf{S}_{+}^\top-\mathbf{S}_{-}^\top] \in\mathbb{R}^{4r\times N}$ and $\bm{B} =[-\bm{I}_{2r};-\bm{I}_{2r}]$.

\subsection{ODKLA: online decentralized kernel learning via linearized ADMM}
To solve \eqref{eq:online_opt_matrix} via ADMM, we first get the augmented Lagrangian form of \eqref{eq:online_opt_matrix} as
\begin{equation}
	\label{eq:aug}
	\begin{split}
		& \textstyle	\mathbb{L}_t(\bm{\Theta},\bm{Z}, \bm{\Lambda})= \mathcal{L}_{t} (\bm{\Theta}) + \frac{\eta_t}{2}\|\bm{\Theta} - \bm{\Theta}_{t}\|_F^2 \\
		&\textstyle  \qquad \qquad +  \left\langle \bm{\Lambda},\bm{A}\bm{\Theta} + \bm{B}\bm{Z} \right\rangle + \frac{\rho}{2}\|\bm{A}\bm{\Theta} + \bm{B}\bm{Z}\|_F^2, 
	\end{split}
\end{equation}
where $\rho$ is the penalty parameter, $\bm{\Lambda} = [\bm{\beta};\bm{\lambda}]\in\mathbb{R}^{4r\times 2L}$ is the Lagrange multiplier associated with the constraint $\bm{A}\bm{\Theta} + \bm{B}\bm{Z} = \mathbf{0}$. Then, at time $t$, the updates of the primal variables $\bm{\Theta}_{t+1}$, $\bm{Z}_{t+1}$ and the dual variable $\bm{\Lambda}_{t+1}$ are sequentially given by 
\begin{align}
	\begin{split}
		\label{eq:odkla_primal}  
		\textstyle	\bm{\Theta}_{t+1}&:=\arg\min_{\bm{\Theta}} \mathbb{L}_t(\bm{\Theta},\bm{Z}_t, \bm{\Lambda}_t),%\\
		%& =\arg\min_{\bm{\Theta}} \mathcal{L}_{t} (\bm{\Theta}) + \frac{\eta_t}{2}\|\bm{\Theta} - \bm{\Theta}_{t}\|_F^2 \\
		%&   \quad + \left\langle \bm{\Lambda}_t,\bm{A}\bm{\Theta} + \bm{B}\bm{Z}_t \right\rangle + \frac{\rho}{2}\|\bm{A}\bm{\Theta} + \bm{B}\bm{Z}_t\|_F^2,
	\end{split}\\
	\begin{split}
		\textstyle	\label{eq:odkla_primal2}
		\bm{Z}_{t+1}&:=\arg\min_{\bm{Z}} \mathbb{L}_t(\bm{\Theta}_{t+1},\bm{Z}, \bm{\Lambda}_t),%\\
		%	&= \arg\min_{\bm{Z}} \left\langle \bm{\Lambda}_{t+1},\bm{A}\bm{\Theta}_{t+1} + \bm{B}\bm{Z} \right\rangle + \frac{\rho}{2}\|\bm{A}\bm{\Theta}_{t+1} + \bm{B}\bm{Z}\|_F^2,
	\end{split} \\
	\begin{split}
		\label{eq:odkla_dual}
		\textstyle  \bm{\Lambda}_{t+1} &= \bm{\Lambda}_{t} + \rho(\bm{A}\bm{\Theta}_{t+1} + \bm{B}\bm{Z}_{t+1}).
	\end{split}
\end{align}
Note that given the instantaneous loss $ \mathcal{L}_{t}$, iterates \eqref{eq:odkla_primal}-\eqref{eq:odkla_dual} only run once, and thus the optimization problem in \eqref{eq:online_opt_matrix} is only approximately solved. It has been proven in \cite{xu2015online} that with initializations $\bm{\beta}_1=-\bm{\lambda}_1$, and $\bm{Z}_1=\frac{1}{2}\mathbf{S}_{+}^\top\bm{\Theta}_{1}$, the update of the auxiliary variable $\bm{Z}_t$ is not necessary and the Lagrange multiplier $\bm{\Lambda}$ can be replaced by a lower dimensional variable $\bm{\Gamma}:=[\bm{\gamma}_1^\top;\cdots;\bm{\gamma}_N^\top ] \in\mathbb{R}^{N\times 2L}$. The simplified updates of ADMM for general online decentralized optimization refer to\cite{xu2015online}. Though simplified, the general decentralized ADMM still involves solving local optimization problem for the primal variables update, thus is computational intensive. To reduce the computation complexity of ADMM, we replace $\mathcal{L}_{t} (\bm{\Theta})$ in \eqref{eq:odkla_primal} by its linear approximation $\mathcal{L}_{t} (\bm{\Theta}_t) + \langle \partial \mathcal{L}_{t} (\bm{\Theta}_{t}), \bm{\Theta} - \bm{\Theta}_{t}\rangle$ at $\bm{\Theta} = \bm{\Theta}_t$, and develop the \textbf{O}nline \textbf{D}ecentralized \textbf{K}ernel learning via \textbf{L}inearized \textbf{A}DMM (ODKLA) algorithm where the iterates of $\bm{\Theta}_{t+1}$ and $\bm{\Gamma}_{t+1}$ are generated by the simplified recursions 
\begin{align}
	\begin{split}
		\label{eq:odkla_simple}
		\textstyle	\bm{\Theta}_{t+1} &=  (\eta_t\mathbf{I} + 2\rho\bm{D})^{-1} \Big[ (\rho(\bm{D}+\bm{W}) + \eta_t\mathbf{I} )\bm{\Theta}_{t} \\
		&\qquad - \bm{\Gamma}_{t} - \partial \mathcal{L}_{t} (\bm{\Theta}_{t})  \Big],
	\end{split}\\
	\begin{split}
		\textstyle		\bm{\Gamma}_{t+1} &= \bm{\Gamma}_{t} + \rho(\bm{D}-\bm{W})\bm{\Theta}_{t+1} .
	\end{split}
\end{align}
The ODKLA algorithm can be implemented distributedly. Specifically, each agent $i$ only needs to update a primal variable $\bm{\theta}_{i}$ and a dual variable $\bm{\gamma}_i$ with the following iterations
\begin{align}
	\begin{split}
		\label{eq:dec_odkla_primal}
		\textstyle	\bm{\theta}_{i,t+1} &= \bm{\theta}_{i,t} - \frac{1}{\eta_{t} + 2\rho d_i}\Big[ \partial \mathcal{L}_{i,t} (\bm{\theta}_{i,t}) \\
		&	\textstyle \qquad +  \rho \sum_{j\in\mathcal{N}_i} (\bm{\theta}_{i,t}- \bm{\theta}_{j,t}) +  \bm{\gamma}_{i,t}  \Big],
	\end{split}\\
	\begin{split}
		\label{eq:dec_odkla_dual}
		\textstyle	\bm{\gamma}_{i,t+1} &= \bm{\gamma}_{i,t} + \rho\sum_{j\in\mathcal{N}_i} (\bm{\theta}_{i,t+1}-\bm{\theta}_{j,t+1}). 
	\end{split}
\end{align}
Note that with linearized ADMM, at each time $t$, ODKLA has closed-form solutions for all agents to update their primal variables, instead of solving optimization problems as in \eqref{eq:odkla_primal}. Thus, the computational efficiency is improved. The ODKLA algorithm is outlined in Algorithm \ref{alg:algo_odkla}.
\begin{algorithm}[t]
	{\small%\fontsize{9pt}{10pt}\selectfont
		\caption{ODKLA (run at agent $i$)}	\label{alg:algo_odkla}
		\begin{algorithmic}[1]
			\REQUIRE Kernel $\kappa$, hyper-parameters $(L, \eta_t)$, initialize local variables to $\bm{\theta}_{i,1}= \mathbf{0}$, and $\bm{\gamma}_{i,1}=0$.
			\STATE Draw $L$ i.i.d. samples $\{\bm{\omega}_l\}_{l=1}^L$ from $p_\kappa(\bm{\omega})$ according to a common random seed.
			\FOR {iterations $t = 1, 2, \dots, T$}
			\STATE Receive a streaming data $(\mathbf{x}_{i,t}, y_{i,t})$
			\STATE Construct $\bm{\phi}(\mathbf{x}_{i,t})$  via \eqref{eq:RF_map}. 
			\STATE Update local primal variable $\bm{\theta}_{i,t+1}$ via \eqref{eq:dec_odkla_primal}.
			\STATE Transmit $\bm{\theta}_{i,t+1}$ to neighbors and receive  $\bm{\theta}_{j,t+1}$ from neighbors $j\in\mathcal{N}_i$.
			\STATE Update local dual variable $\bm{\gamma}_{i,t+1}$ via \eqref{eq:dec_odkla_dual}.          
			\ENDFOR
	\end{algorithmic}}
\end{algorithm} 

\textbf{Remark 1}. Our paper shares similar problem formulation \eqref{eq:online_opt} as \cite{hong2021distributed}. However, our methods differ from~\cite{hong2021distributed} in two ways. First, we utilize linearized ADMM to solve the decentralized kernel learning problem while \cite{hong2021distributed} adopts the standard ADMM. Compared with~\cite{hong2021distributed}, our algorithms enjoy light computation. Second, we also develop the communication efficient algorithms in the next section using quantization and communication censoring strategies while the communication efficiency is not discussed in  \cite{hong2021distributed}.

\subsection{QC-ODKLA: quantized and communication-censored ODKLA}

ODKLA resolves the challenges caused by streaming data in decentralized network setting in a computationally efficient manner. However, as seen in \eqref{eq:dec_odkla_primal} - \eqref{eq:dec_odkla_dual}, agents communicate all the time which causes low communication efficiency. Thus, we introduce communication censoring and quantization strategies to deal with the limited communication resource situation and develop the \textbf{Q}uantized and  \textbf{C}ommunication-censored \textbf{ODKLA} algorithm (QC-ODKLA).

To start, we introduce a new state variable $\hat{\bm{\theta}}_{i,t}$ for each agent $i$ to record its latest broadcast primal variable up to time $t$. Then, the difference between agent $i$'s updated state $\bm{\theta}_{i,t+1}$ and its previously transmitted state $\hat{\bm{\theta}}_{i,t}$ at time $t$ is defined as
\begin{equation}
	\label{eq:diff_h}
	\textstyle		\bm{h}_{i,t} = \bm{\theta}_{i,t+1} - \hat{\bm{\theta}}_{i,t}.
\end{equation}

We then introduce an evaluation function 
\begin{equation}
	\label{eq:censorfuc}
	\textstyle H_{i,t} = \|\bm{h}_{i,t}\|_2 - \alpha\beta^t 
\end{equation}
to evaluate if the local updates $\bm{\theta}_{i,t+1}$ are informative enough to be transmitted, with predefined positive constants $\alpha>0$ and $\beta<1$. If $H_{i,t}\geq 0$, then $\bm{\theta}_{i,t+1}$ is deemed informative, and agent $i$ is allowed to transmit a quantized update $Q(\bm{\theta}_{i,t+1})$ to its neighbors. Here, the quantization is introduced to reduce the communication cost from the perspective of bit numbers per transmission. To facilitate the measurement and analysis of the impact of quantization, we adopt the difference-based quantization scheme proposed in~\cite{liu2021dqc}. That is, at time $t$, instead of quantizing $\bm{\theta}_{i,t+1}$, we quantize the difference $\bm{h}_{i,t}$. Specifically, for each element $h_{i,t}^l, l=1,\dots, 2L$ within the range of $[u,v)$, if we restrict the number of transmission bits to be $b$, then we can evenly divided the range $[u,v)$ to be $q=2^b$ intervals of equal length $\Delta = (v-u)/q$. Then the rounding quantizer $Q(\cdot)$ applied to $h_{i,t}^l$ outputs
\begin{equation}
	\label{eq:quantizer}
	\textstyle	Q(h_{i,t}^l) =  u+ \left(  \floor*{\frac{h_{i,t}^l-u}{\Delta}} + \frac{1}{2} \right) \Delta,
\end{equation}
where $\floor*{\cdot}$ is the floor operation. In practice, it is not necessary to transmit $Q(h_{i,t}^l)$, instead, we can simply transmit the integer $k:=\floor*{\frac{h_{i,t}^l-u}{\Delta}}$ using the $b$ bits. Thus, the total number of bits for agent $i$ to transmit the quantized difference $Q(\bm{h}_{i,t})$ to its neighbors is only $2Lb$ bits.  

The whole communication process thus involves three parts, evaluation, quantization, and state update. If $H_{i,t}\geq 0$, then $\bm{\theta}_{i,t+1}$ is deemed informative, and agent $i$ is allowed to transmit a quantized difference $Q(\bm{h}_{i,t})$ to its neighbors and updates its local state as $\hat{\bm{\theta}}_{i,t+1} =\hat{\bm{\theta}}_{i,t} + Q(\bm{h}_{i,t})$. Otherwise, $\bm{\theta}_{i,t+1}$ is censored, agent $i$ sets $\hat{\bm{\theta}}_{i,t+1}=\hat{\bm{\theta}}_{i,t}$, and no information is transmitted. Similarly, upon receiving $Q(\bm{h}_{j,t})$ from its neighbor $j$, agent $i$ updates the state variables of its neighbor's as $\hat{\bm{\theta}}_{j,t+1} =\hat{\bm{\theta}}_{j,t} + Q(\bm{h}_{j,t})$, otherwise, $\hat{\bm{\theta}}_{j,t+1}=\hat{\bm{\theta}}_{j,t}$. 

With the communication censoring rule and quantization scheme, the primal and dual updates in \eqref{eq:dec_odkla_primal} and \eqref{eq:dec_odkla_dual} become
\begin{align}
	\begin{split}
		\label{eq:dec_qcodkla_primal}
		\textstyle		\bm{\theta}_{i,t+1} &= \bm{\theta}_{i,t} - \frac{1}{\eta_{t} + 2\rho d_i}\Big[ \partial \mathcal{L}_{i,t} (\bm{\theta}_{i,t})  \\
		\textstyle	\qquad & +  \rho \sum_{j\in\mathcal{N}_i} (\hat{\bm{\theta}}_{i,t} - \hat{\bm{\theta}}_{j,t}) +  \bm{\gamma}_{i,t}  \Big] ,
	\end{split}\\
	\begin{split}
		\label{eq:dec_qcodkla_dual}
		\textstyle		\bm{\gamma}_{i,t+1} &= \bm{\gamma}_{i,t} + \rho\sum_{j\in\mathcal{N}_i}\Big(\hat{\bm{\theta}}_{i,t+1}-\hat{\bm{\theta}}_{j,t+1}\Big),
	\end{split}
\end{align}
and the total numbers of transmissions and bits are both reduced in the optimization and learning process. We summarize the QC-ODKLA algorithm in Algorithm 2. 

\begin{algorithm}[t]
	{\small%\fontsize{9pt}{10pt}\selectfont
		\caption{QC-ODKLA (run at agent $i$ )}\label{alg:algo_qcodkla}
		\begin{algorithmic}[1]
			\REQUIRE Kernel $\kappa$, hyper-parameters $(L, \rho, \alpha, \beta)$, initialize local variables to $\bm{\theta}_{i,1}= \mathbf{0}$, and $\bm{\gamma}_{i,1}=\mathbf{0}$, $\hat{\bm{\theta}}_{i,1}= Q(\bm{\theta}_{i,1})$ and $\hat{\bm{\theta}}_{j,1}= Q(\bm{\theta}_{j,1})$ for all $j\in\mathcal{N}_i$. 
			\STATE Draw $L$ i.i.d. samples $\{\bm{\omega}_l\}_{l=1}^L$ from $p_\kappa(\bm{\omega})$ according to a common random seed.
			\FOR {iterations $t = 1, 2, \dots, T$}
			\STATE Receive a streaming data $(\mathbf{x}_{i,t}, y_{i,t})$
			\STATE Construct $\bm{\phi}(\mathbf{x}_{i,t})$  via \eqref{eq:RF_map}. 
			\STATE Update local primal variable $\bm{\theta}_{i,t+1}$ by solving \eqref{eq:dec_qcodkla_primal}.
			\STATE Calculate the difference $\bm{h}_{i,t}$ via \eqref{eq:diff_h} and quantize it as $Q(\bm{h}_{i,t})$ via \eqref{eq:quantizer}.
			\STATE If \eqref{eq:censorfuc} is nonnegative, transmit $Q(\bm{h}_{i,t})$ to neighbors and set $\hat{\bm{\theta}}_{i,t+1} =\hat{\bm{\theta}}_{i,t} + Q(\bm{h}_{i,t})$. Else, set $\hat{\bm{\theta}}_{i,t+1} =\hat{\bm{\theta}}_{i,t}$ and do not transmit.
			\STATE If receiving $Q(\bm{h}_{j,t})$ from neighbors $j$, update  $\hat{\bm{\theta}}_{j,t+1} =\hat{\bm{\theta}}_{j,t} + Q(\bm{h}_{j,t})$. Else, set $\hat{\bm{\theta}}_{j,t+1} =\hat{\bm{\theta}}_{j,t}$.
			\STATE Update local dual variable $\bm{\gamma}_{i,t+1}$ via \eqref{eq:dec_qcodkla_dual}.          
			\ENDFOR
	\end{algorithmic}}
\end{algorithm}

\section{Regret Analysis}
\label{sec:regret}
In this section, we analyze the regret bound of QC-ODKLA. As in~\cite{bouboulis2018online}, we define the cumulative network regret of online decentralized learning as
\begin{equation}
\label{eq:stat_reg}
\begin{split}
\textstyle		\mathcal{R}(T) &=  \sum_{t=1}^T  \sum_{i=1}^N \left( \mathcal{L}_{i,t}(\bm{\theta}_{i,t}) -  \mathcal{L}_{i,t}(\bm{\theta}^\star)\right) 
\end{split}
\end{equation}
where $\bm{\theta}^\star$ is the optimal solution of \eqref{eq:decentr_theta} that assumes all data are available. We prove that QC-ODKLA achieves the optimal sublinear regret $\mathcal{O}(\sqrt{T})$ for convex local cost functions $\mathcal{L}_{i,t}$. Since ODKLA is a special case of QC-ODKLA where both the quantization and communication-censoring strategies are absent, the regret analysis of QC-ODKLA extends to ODKLA straightforwardly. The following commonly used assumptions are adopted. 

%\begin{assumption}
%	\label{ass:net_connect}
%	The network with topology $\mathcal{G}=(\mathcal{N}, \mathcal{A})$ is undirected and connected.
%\end{assumption}

\begin{assumption}
	\label{ass:convex}
	The local cost functions $\mathcal{L}_{i,t}(\bm{\theta})$ are convex and differentiable with respect to $\bm{\theta}$. Also, assume the gradients of the local cost functions are Lipschitz continuous with constants $C_{\mathcal{L}_i}>0, \forall\; i$. That is, $\|\partial\mathcal{L}_{i,t}(\bm{\theta})\|_2\leq  C_{\mathcal{L}_i}, \forall\; i$. The maximum Lipschitz constant is $C_{\mathcal{L}}:=\max_{i} C_{\mathcal{L}_i}$.
\end{assumption}

\begin{assumption}
	\label{ass:boundtheta}
	The estimates $\bm{\theta}_{i,t}$ and the optimal solution $\bm{\theta}^\star$ of \eqref{eq:decentr_theta} are bounded. That is, $\|\bm{\theta}_{i,t}\|_2\leq C_{\theta}$, and $\|\bm{\theta}^\star\|_2\leq C_{\theta}$. 
\end{assumption}

Note that all assumptions are standard in online decentralized kernel learning~\cite{bouboulis2018online, hong2021distributed, shen2021distributed}. The convexity of local cost functions are easily satisfied in learning problems if the local cost functions are square loss or the hinge loss. 
%which implies $\|\bm{\Theta}^\star\|_F\leq \sqrt{N}C_{\theta}$

To study the regret bound for QC-ODKLA, we notice that the difference of QC-ODKLA and ODKLA is the communication censoring step and quantization step in the communication stage, which introduces an error if an update is censored and/or quantized in an transmission. We define the introduced error for agent $i$ at time $t$ as  
\begin{equation}
	\label{eq:intro_error_i}
	\bm{e}_{i,t}: = \bm{\theta}_{i,t} - \hat{\bm{\theta}}_{i,t},
\end{equation} 
and the overall introduced error at time $t$ as $\bm{E}_t: =[\bm{e}_{1,t}^\top; \bm{e}_{2,t}^\top;\dots;\bm{e}_{N,t}^\top]$. We first show that the overall introduced error in QC-ODKLA is upper bounded by the quantization error and the pre-defined threshold parameters.
\begin{lemma}
	\label{lem:lemma1}
	For the updates \eqref{eq:dec_qcodkla_primal} and \eqref{eq:dec_qcodkla_dual}, under the assumptions \ref{ass:convex} and \ref{ass:boundtheta}, if the quantized difference $Q(\bm{h}_{i,t})$ is only allowed to transmit when $H_{i,t} \geq 0$ for the pre-defined threshold parameters $\alpha$ and $\beta$, then, for any time $t>0$, the overall error introduced in the QC-ODKLA is upper bounded by
	\begin{equation}
		\label{eq:bounded_error}
		\|\bm{E}_t\|_F^2 \leq \zeta:=\max\{\sqrt{N}\alpha\beta, \sqrt{2NL}\Delta/2\},
	\end{equation}
	where $\Delta$ is the length of the quantization interval.
\end{lemma} 

\textit{Proof.} Define $\delta \hat{\bm{\theta}}_{i,t} = \hat{\bm{\theta}}_{i,t} - \hat{\bm{\theta}}_{i,t-1}$, the introduced error for each agent $i$ can be represented as
\begin{equation}
\label{eq:intro_error_i_resp}
\begin{split}
\bm{e}_{i,t}  &= \bm{\theta}_{i,t} - \hat{\bm{\theta}}_{i,t} \\
&= \bm{\theta}_{i,t} - \hat{\bm{\theta}}_{i,t-1} - \delta \hat{\bm{\theta}}_{i,t} \\
&= \bm{h}_{i,t-1} - \delta \hat{\bm{\theta}}_{i,t}.
\end{split}
\end{equation}
According to the censoring rule, if $\| \bm{h}_{i,t-1}\|_2 \geq \alpha \beta^{t-1}$ for $t\geq 1$, we have $\delta \hat{\bm{\theta}}_{i,t} = Q(\bm{h}_{i,t-1})$, which implies $\| \bm{e}_{i,t} \|_2 =\|\bm{h}_{i,t-1} - Q(\bm{h}_{i,t-1})  \|_2 \leq  \sqrt{2L}\Delta/2$. Otherwise, if $\| \bm{h}_{i,t-1}\|_2 < \alpha \beta^{t-1}$ for $t\geq 1$, we have $\delta \hat{\bm{\theta}}_{i,t} = \mathbf{0}$, which implies  $\| \bm{e}_{i,t} \|_2 =\|\bm{h}_{i,t-1} \|_2 \leq  \alpha \beta^{t-1} \leq  \alpha \beta$ since $\beta < 1$. Therefore, the overall introduced error $\|\bm{E}_t\|_F^2 \leq \max\{\sqrt{N}\alpha\beta, \sqrt{2NL}\Delta/2\}$.  \hfill $\blacksquare$  

With Lemma \ref{lem:lemma1}, we are ready to establish the network regret bound of QC-ODKLA.
\begin{theorem}
	\label{theor:theorm2}
	Under the assumptions \ref{ass:convex} - \ref{ass:boundtheta}, if the quantized difference $Q(\bm{h}_{i,t})$ is only allowed to transfer when $H_{i,t} \geq 0$ for the pre-defined threshold parameters $\alpha>0$ and $\beta<1$, then, for any time $t>0$, the cumulative network regret \eqref{eq:stat_reg} generated by the updates \eqref{eq:dec_qcodkla_primal} and \eqref{eq:dec_qcodkla_dual} satisfies
	\begin{equation}
		\label{eq:cum_qcodkla}
		\mathcal{R}(T) \leq(\sqrt{N}C_{\theta} + \frac{1}{\sigma^2_{\max}(\mathbf{S}_{-})}  C_{\mathcal{L}} + \sigma^2_{\max}(\mathbf{S}_{-})\zeta)\mathcal{O}(\sqrt{T})
	\end{equation}
	if $\eta_t = \rho = 1/\mathcal{O} (\sqrt{T})$.
\end{theorem}
\textit{Proof.} See Appendix \ref{subsec:theorem2}. \hfill $\blacksquare$

\textbf{Remark 2.} Note that in addition to the network size ($N$) and topology ($\mathbf{S}_{-}$), the communication censoring and quantization strategies (incorporated in $\zeta$) also affect the cumulative network regret, which creates a trade-off between the communication efficiency and the online learning performance.

\section{Experiments}
\label{sec:exper}
This section evaluates the performance of our ODKLA and QC-ODKLA algorithms in regression tasks for streaming data from real-world datasets.

\textbf{Benchmarks.} Since we consider the case that data are only locally available and cannot be shared among agents, the RFF-DOKL algorithm which is developed based on online gradient descent and a diffusion strategy~\cite{bouboulis2018online} and the DOKL algorithm which is developed based on online ADMM~\cite{hong2021distributed} will be simulated and compared in our experiments with the proposed ODKLA and QC-ODKLA algorithms.  

\textbf{Datasets.} The regression tasks are carried out on 6 datasets available at the UCI machine learning repository~\cite{asuncion2007uci}. The detailed descriptions of the six datasets are listed below.

\noindent
\textbf{Tom's hardware.} This dataset contains $T_{total} = 11000$ samples with $\mathbf{x}_t\in \mathbb{R}^{96}$ including the number of created discussions and authors interacting of a topic and $y_t\in \mathbb{R}$ representing the average number of display to a visitor about that topic~\cite{kawala2013predictions}.  

\noindent
\textbf{Twitter.} This dataset consists of $T_{total} = 98700$ samples with $\mathbf{x}_t\in \mathbb{R}^{77}$ being a feature vector reflecting the number of new interactive authors and the lengths of discussions on a given topic, etc., and $y_t\in \mathbb{R}$ representing the average number of active discussions on a certain topic. The learning task is to predict the popularity of these topics~\cite{kawala2013predictions}. 

\noindent
\textbf{Energy.} This dataset contains $T_{total} = 18600$ samples with $\mathbf{x}_t\in \mathbb{R}^{28}$ describing the
humidity and temperature in different areas of the house, pressure, wind speed, and viability outside, while $y_t$ denotes the total energy consumption in the house~\cite{candanedo2017data}. 

\noindent
\textbf{Air quality.} This dataset contains $T_{total} = 7320$ samples measured by a gas multi-sensor device in an Italian city, where $\mathbf{x}_t\in \mathbb{R}^{13}$ represents the hourly concentration of CO, NOx, NO2, etc, while $y_t$ denotes the concentration of polluting chemicals in the air~\cite{de2008field}.

\noindent
\textbf{Conductivity.} This dataset contains $T_{total} = 21260$ samples extracted from superconductors, where $\mathbf{x}_t\in \mathbb{R}^{81}$ represents critical information to construct  superconductor such as density and mass of atoms. The task is to predict the critical temperature which creates superconductor~\cite{hamidieh2018data}.

\noindent
\textbf{Blood data.} This dataset contains $T_{total} = 61000$ samples recorded by patient monitors at different hospitals where $\mathbf{x}_t\in \mathbb{R}^{2}$ and the goal is to predict the blood pressure based on several physiological parameters from Photoplethysmography and Electrocardiogram signals~\cite{kachuee2015cuff}. 

\textbf{Settings and parameter tuning.} All experiments are conducted using Matlab 2021 on an Intel CPU @ 3.6 GHz (32 GB RAM) desktop. For each dataset, the $T_{total}$ data samples are randomly shuffled and then partitioned among $N$ nodes so that each node has $T = T_{total}/N$ samples. The features are normalized so that all values are between $0$ and $1$. The number of random features adopted for RF approximation is $L = 50$ throughout the simulations. The Gaussian kernel bandwidth is fined tuned to be $\sigma = 0.5$ for Tom's hardware, Twitter, Air quality, and Blood datasets. For Conductivity and Energy datasets, $\sigma = 1$ and  $\sigma = 0.1$, respectively. The regularization parameter $\lambda = 1e-4$. The stepsize $\rho$ and $\eta_t$ are  fine-tuned via grid-search for each method and each dataset individually. The connected graph is randomly generated with $N=5$ or $N=10$ nodes. For Twitter, Conductivity, and Blood datasets, we use a 10-node network. The remaining datasets use a 5-node network. The censoring threshold parameters are $\alpha = 2, \beta =0.9$ for energy data, and  $\alpha= 4, \beta = 0.99$ for all the other datasets. 

\begin{figure*}[!t]
\centering
\subfloat[]{\includegraphics[width=2.2in]{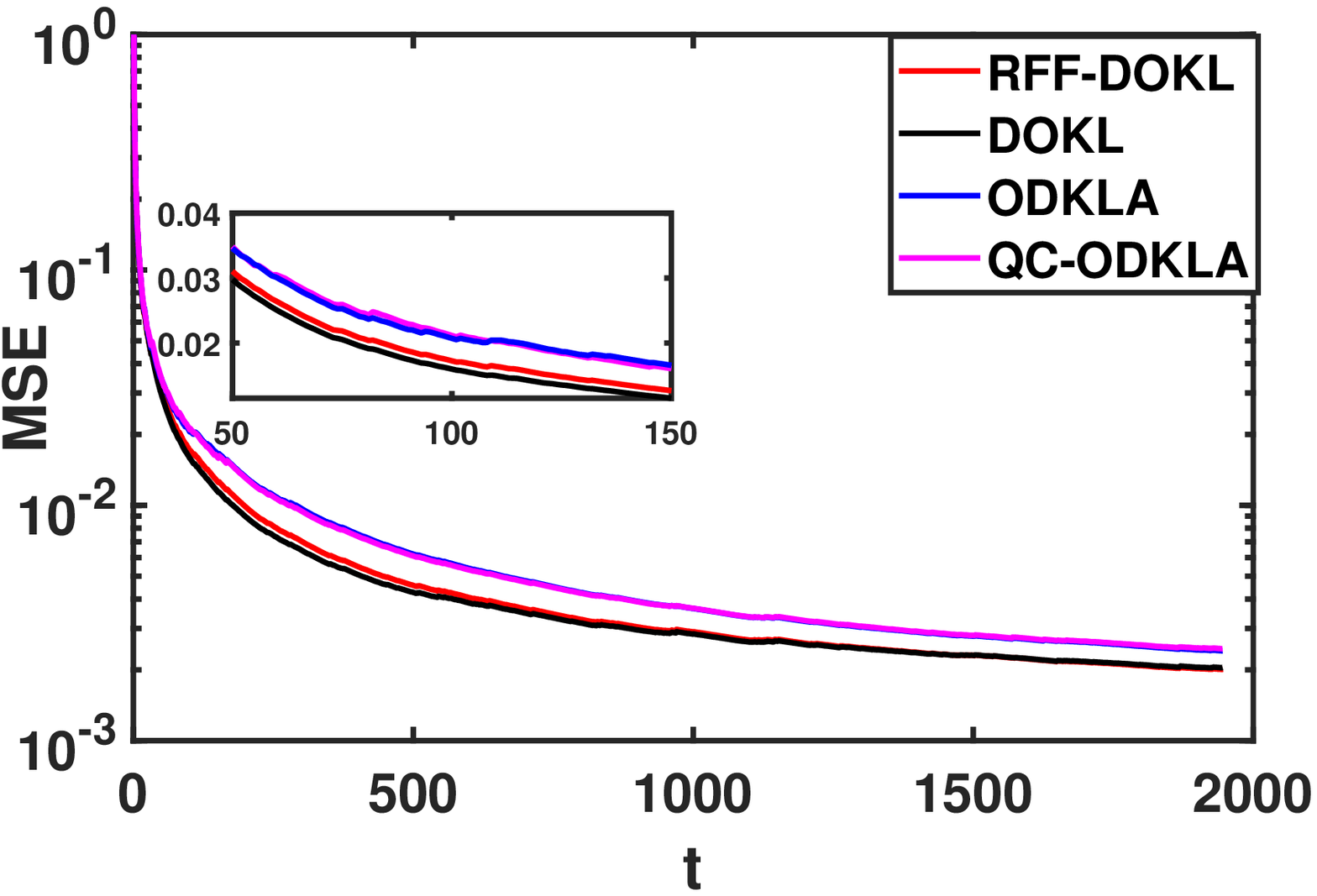}}
\hfil
\subfloat[]{\includegraphics[width=2.2in]{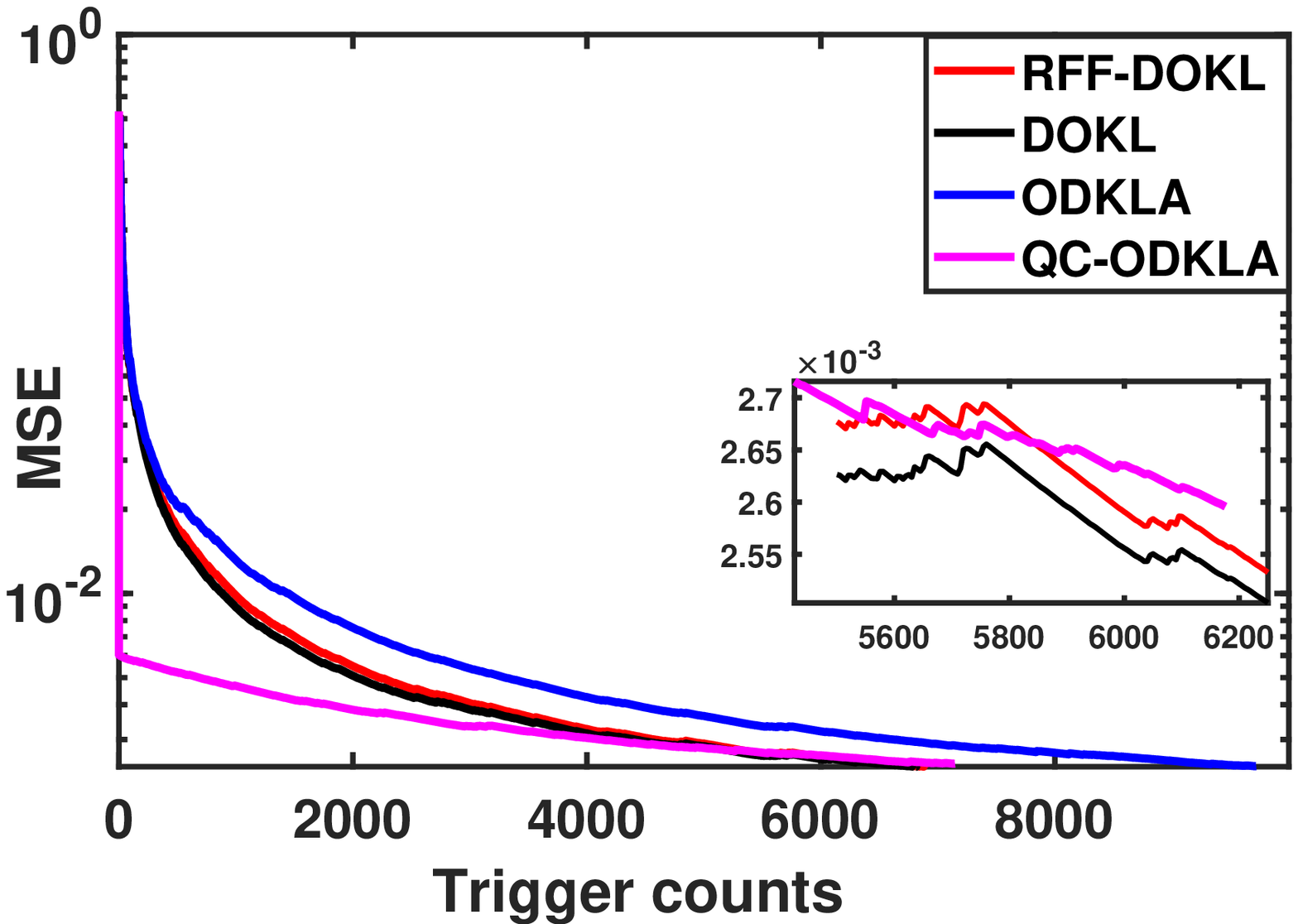}}
\hfil
\subfloat[]{\includegraphics[width=2.2in]{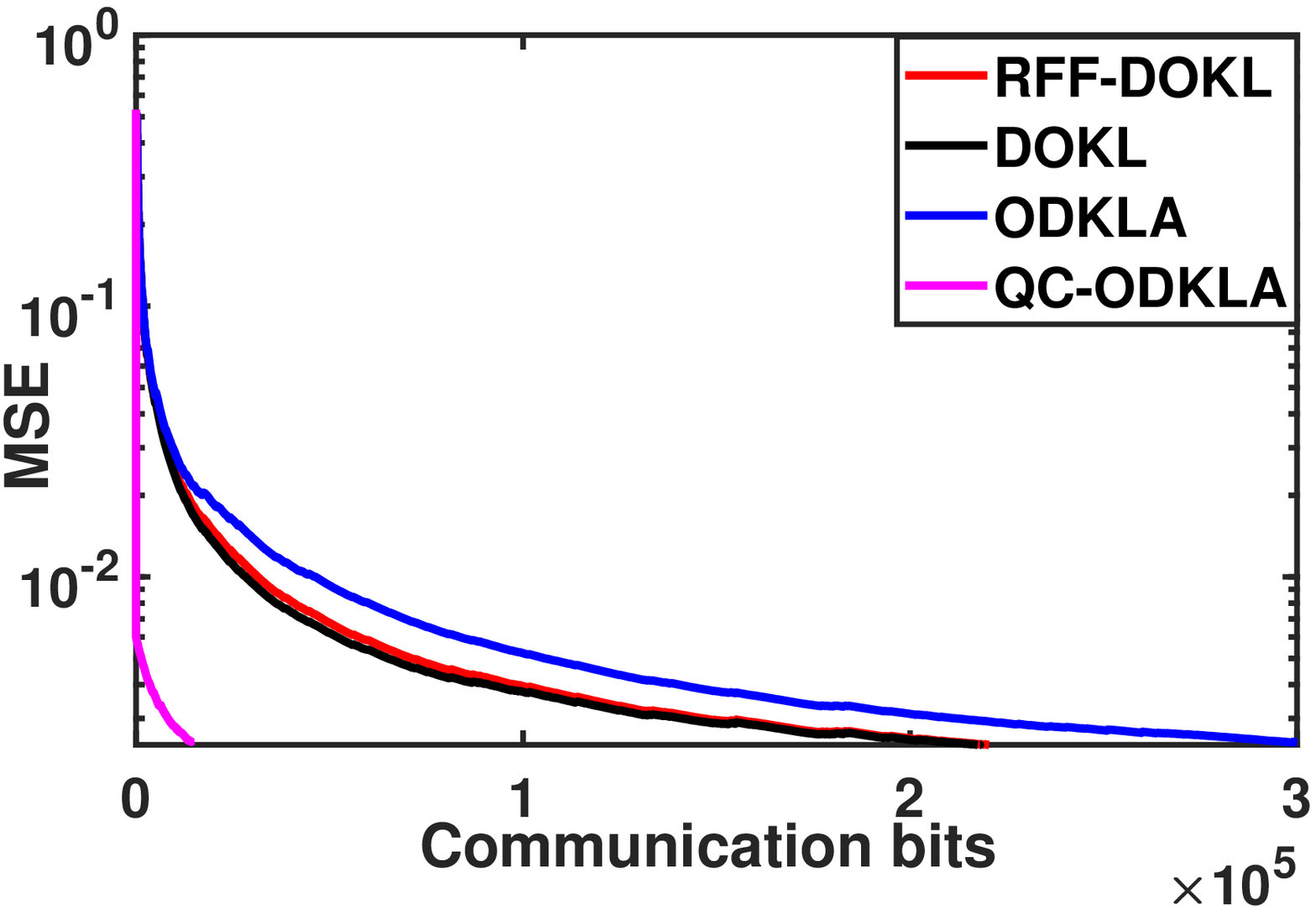}}
\caption{Learning performance on Tom's hardware dataset. (a) MSE vs. time; (b) MSE vs. triggers; (c) MSE vs. bits.}\label{fig:tom}
\end{figure*}

\begin{figure*}[!t]
	\centering
	\subfloat[]{\includegraphics[width=2.2in]{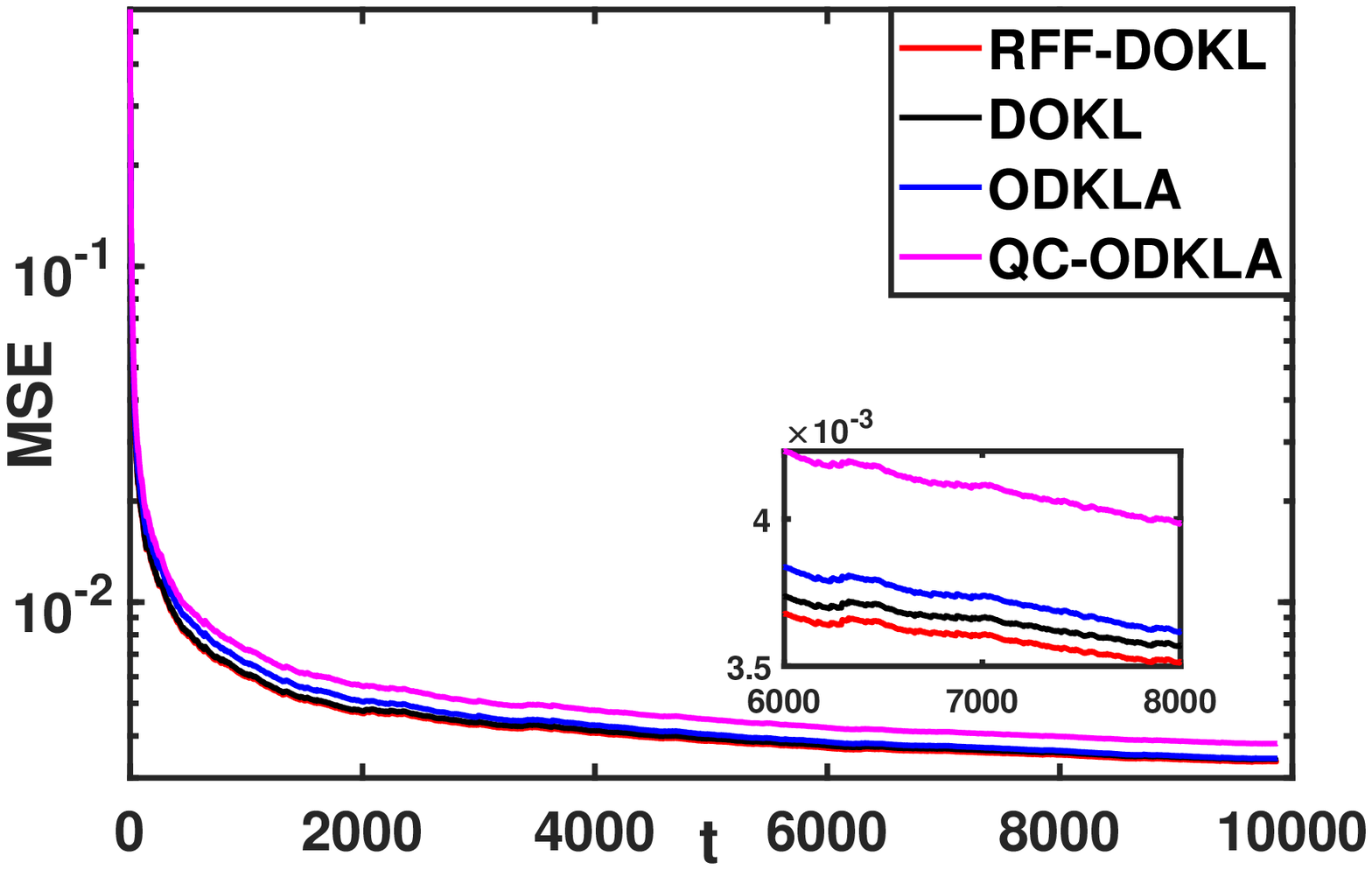}}
	\hfil
	\subfloat[]{\includegraphics[width=2.2in]{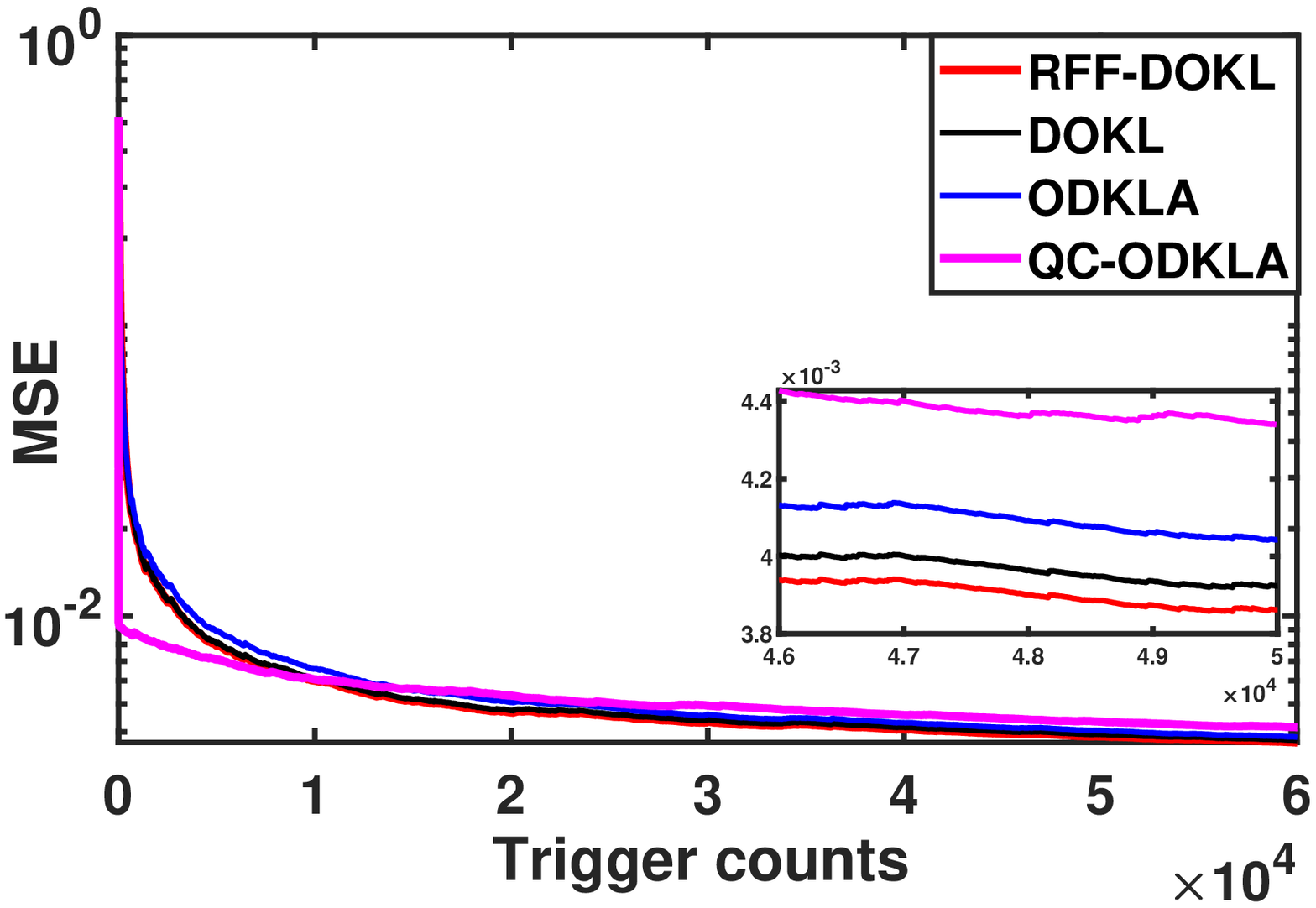}}
	\hfil
	\subfloat[]{\includegraphics[width=2.2in]{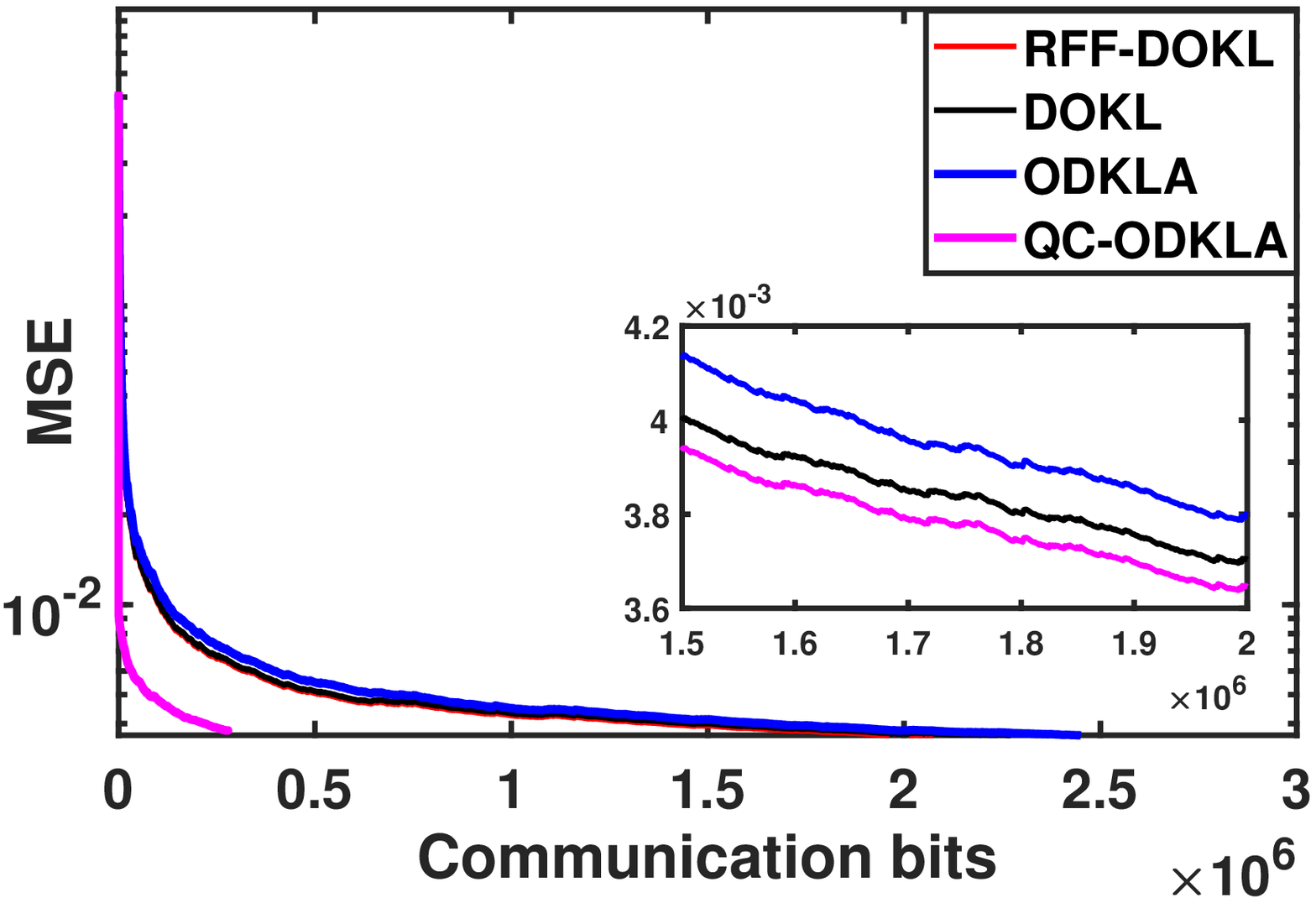}}
	\caption{Learning performance on Twitter dataset. (a) MSE vs. time; (b) MSE vs. triggers; (c) MSE vs. bits.}\label{fig:twitter}
\end{figure*}

\begin{figure*}[!t]
	\centering
	\subfloat[]{\includegraphics[width=2.2in]{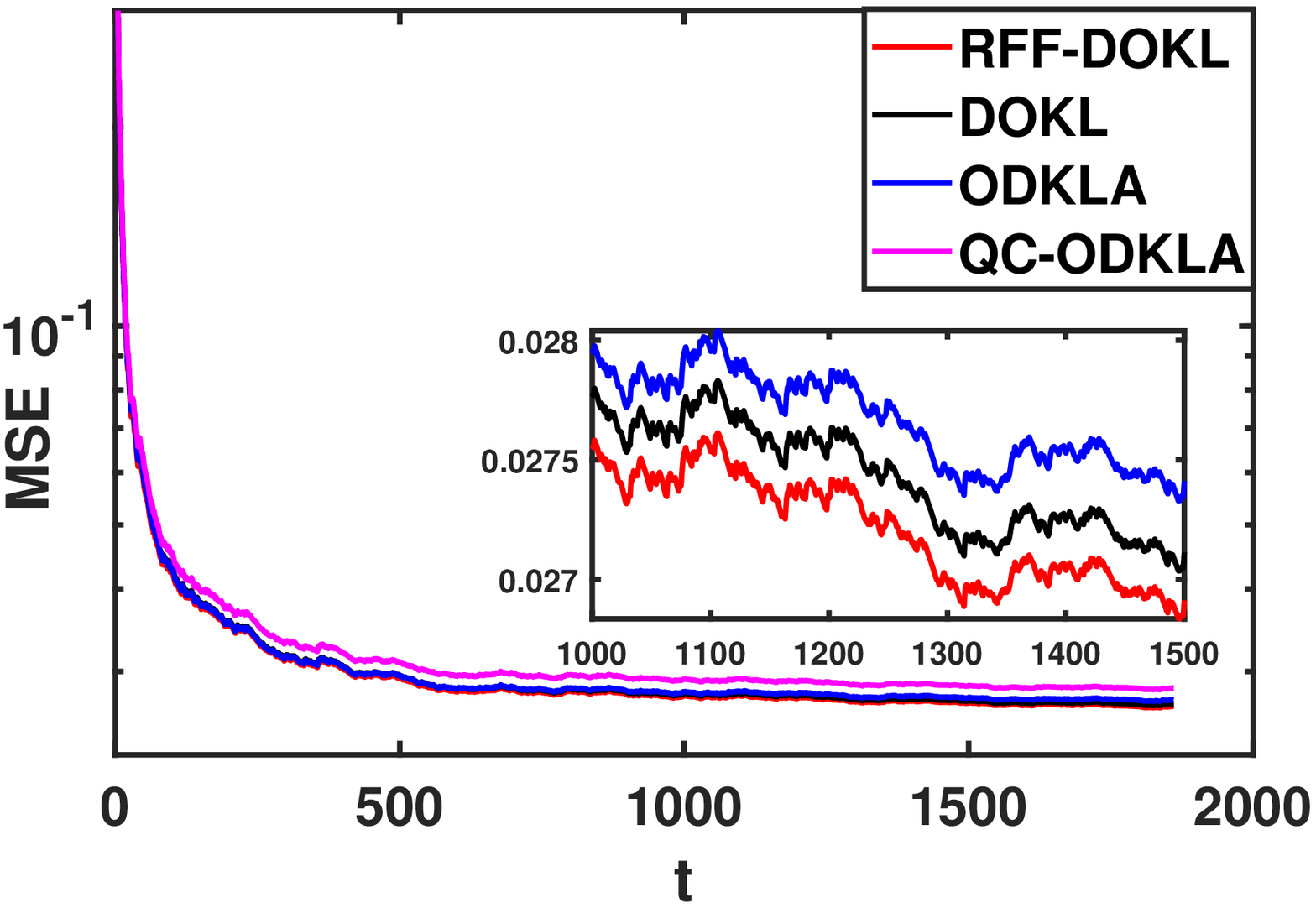}}
	\hfil
	\subfloat[]{\includegraphics[width=2.2in]{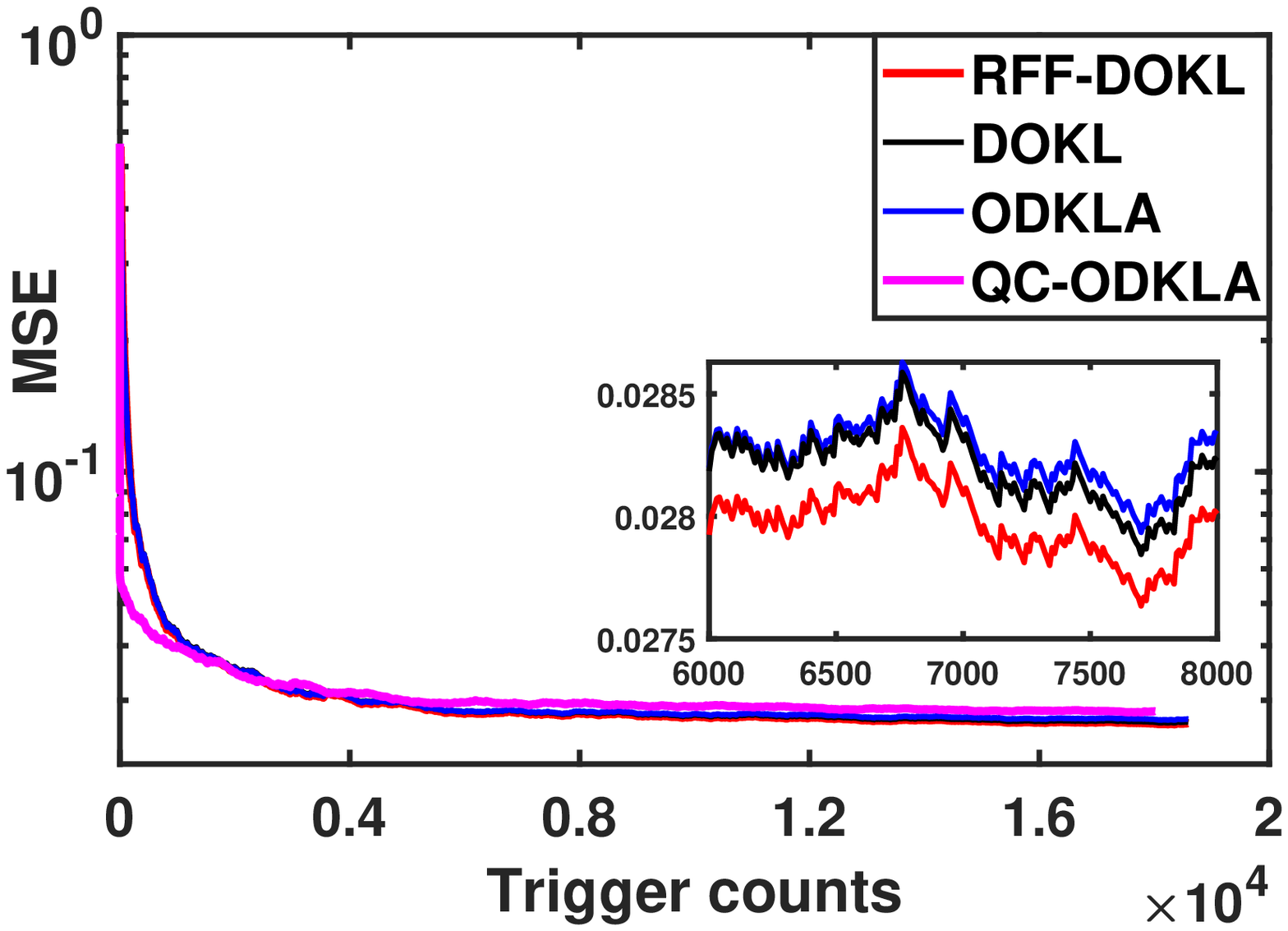}}
	\hfil
	\subfloat[]{\includegraphics[width=2.2in]{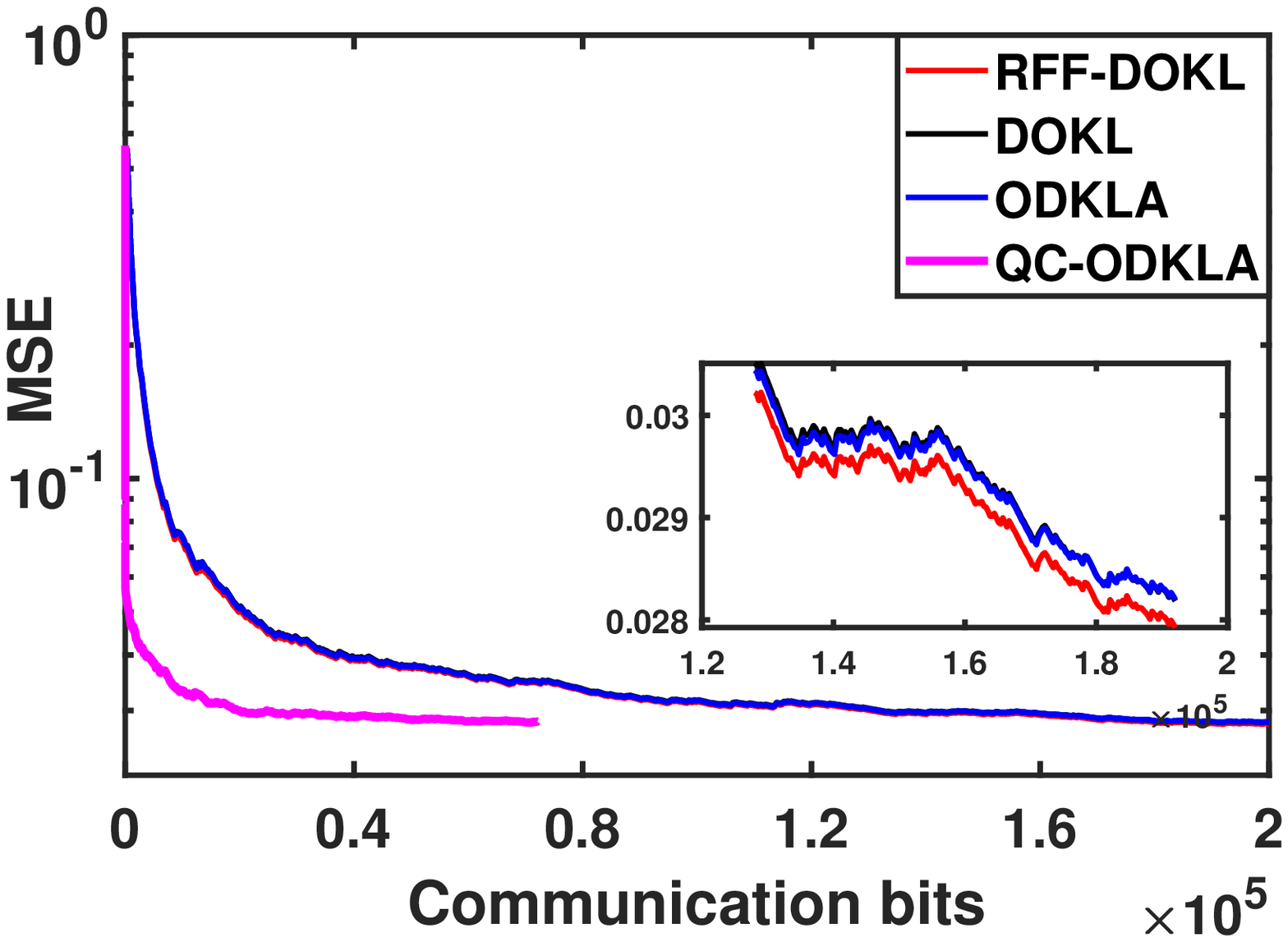}}
	\caption{Learning performance on Energy dataset. (a) MSE vs. time; (b) MSE vs. triggers; (c) MSE vs. bits.}\label{fig:energy}
\end{figure*}

\begin{figure*}[!t]
	\centering
	\subfloat[]{\includegraphics[width=2.2in]{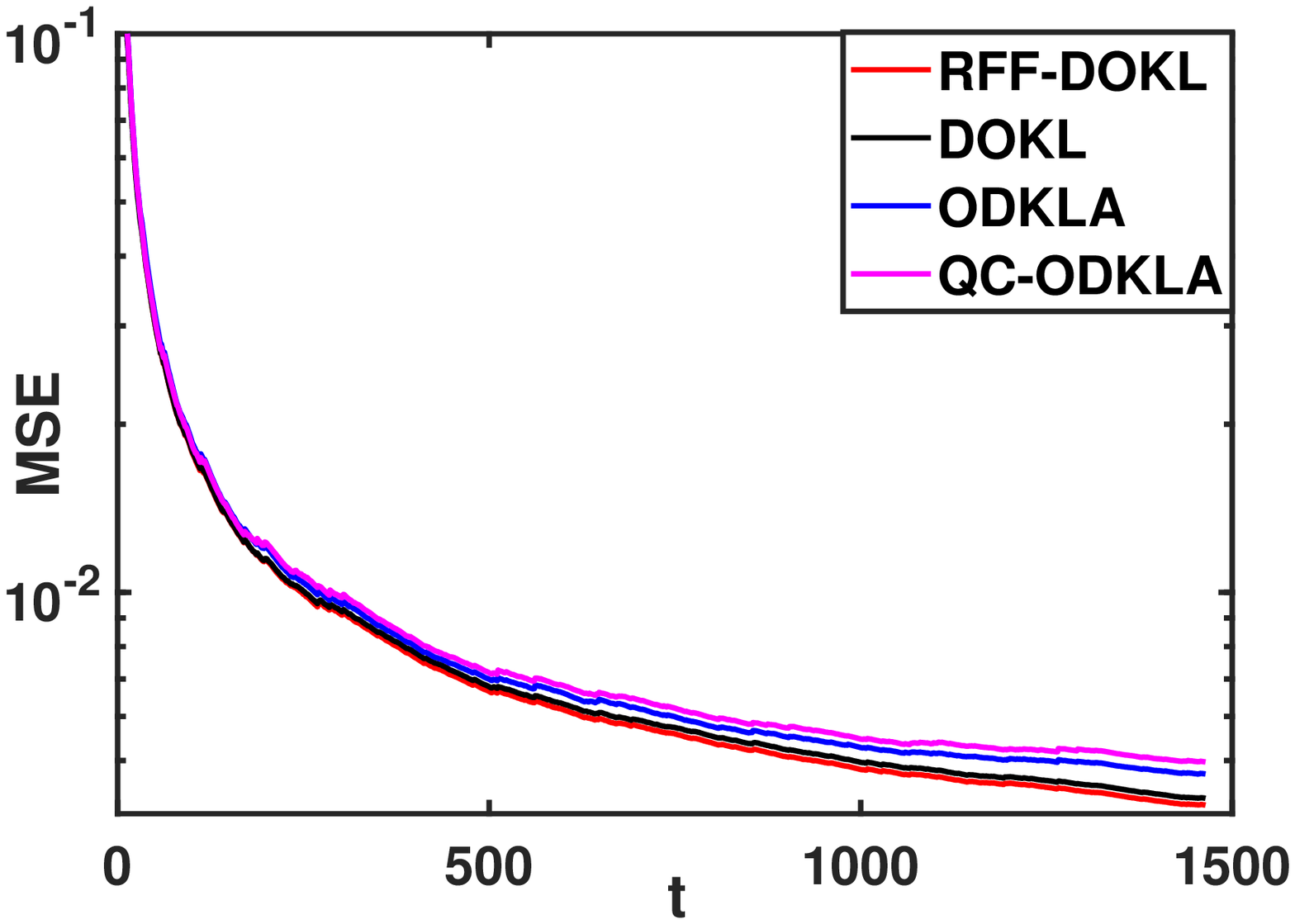}}
	\hfil
	\subfloat[]{\includegraphics[width=2.2in]{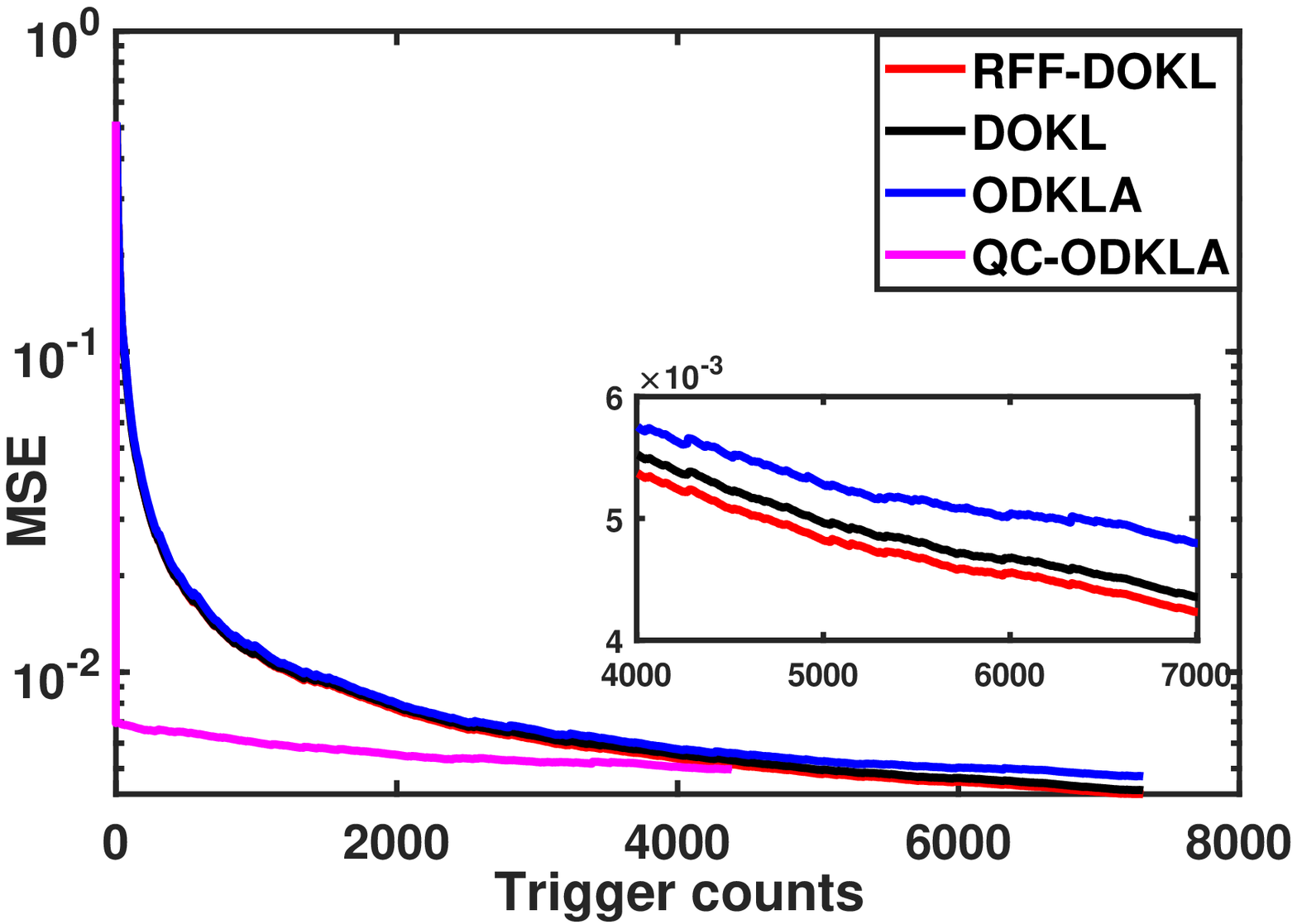}}
	\hfil
	\subfloat[]{\includegraphics[width=2.2in]{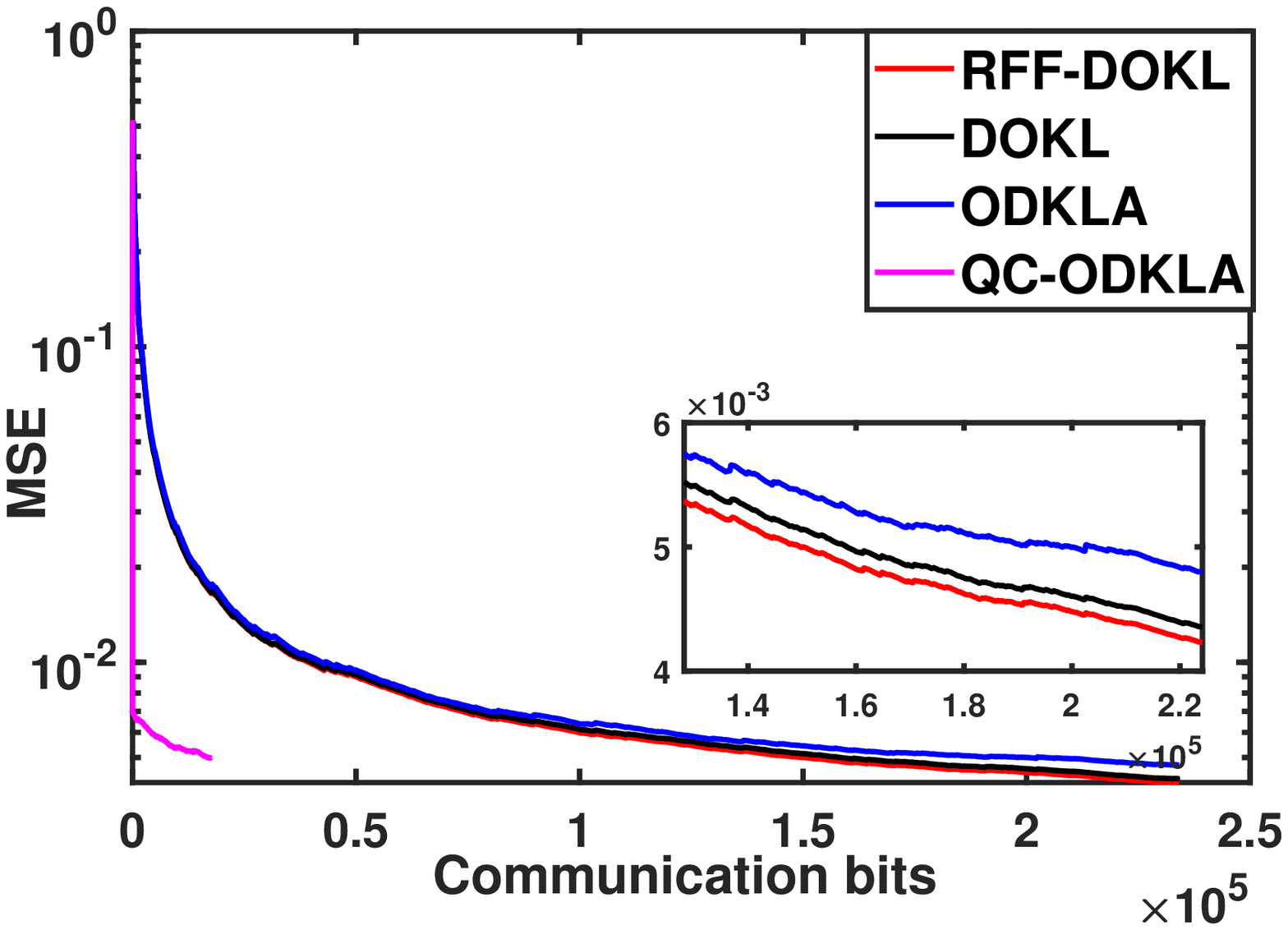}}
	\caption{Learning performance on Air quality dataset. (a) MSE vs. time; (b) MSE vs. triggers; (c) MSE vs. bits.}\label{fig:air}
\end{figure*}

\begin{figure*}[!t]
	\centering
	\subfloat[]{\includegraphics[width=2.2in]{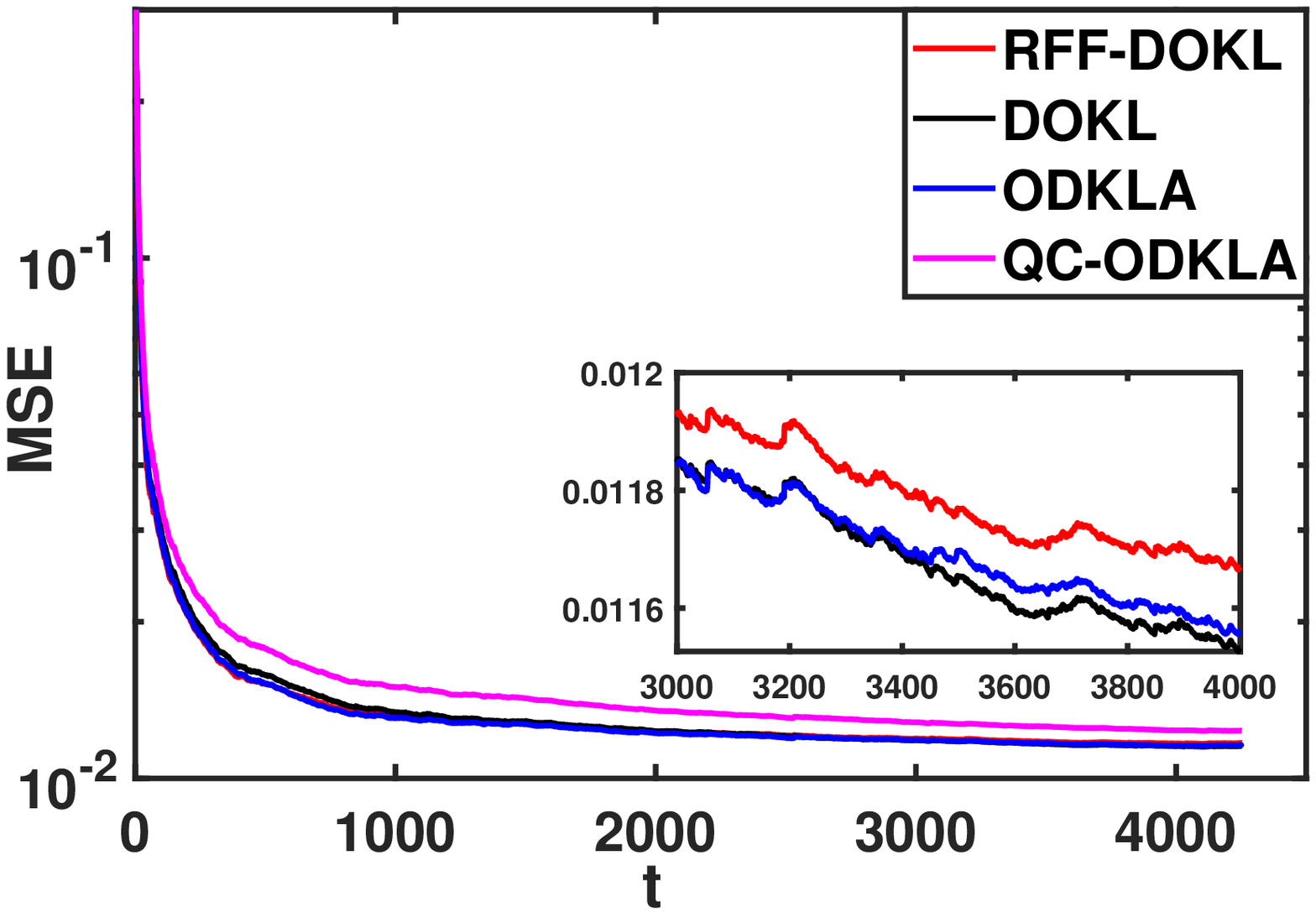}}
	\hfil
	\subfloat[]{\includegraphics[width=2.2in]{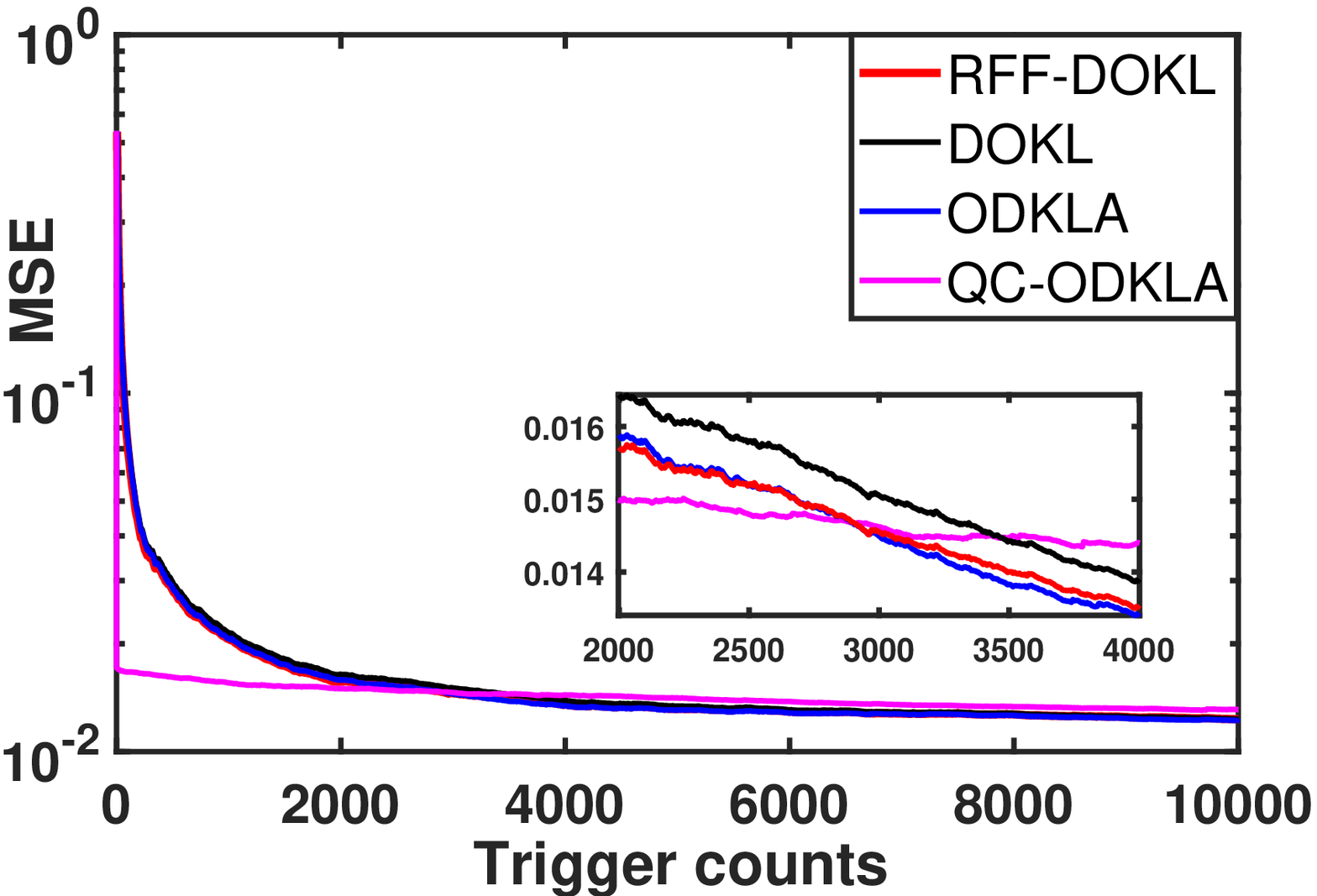}}
	\hfil
	\subfloat[]{\includegraphics[width=2.2in]{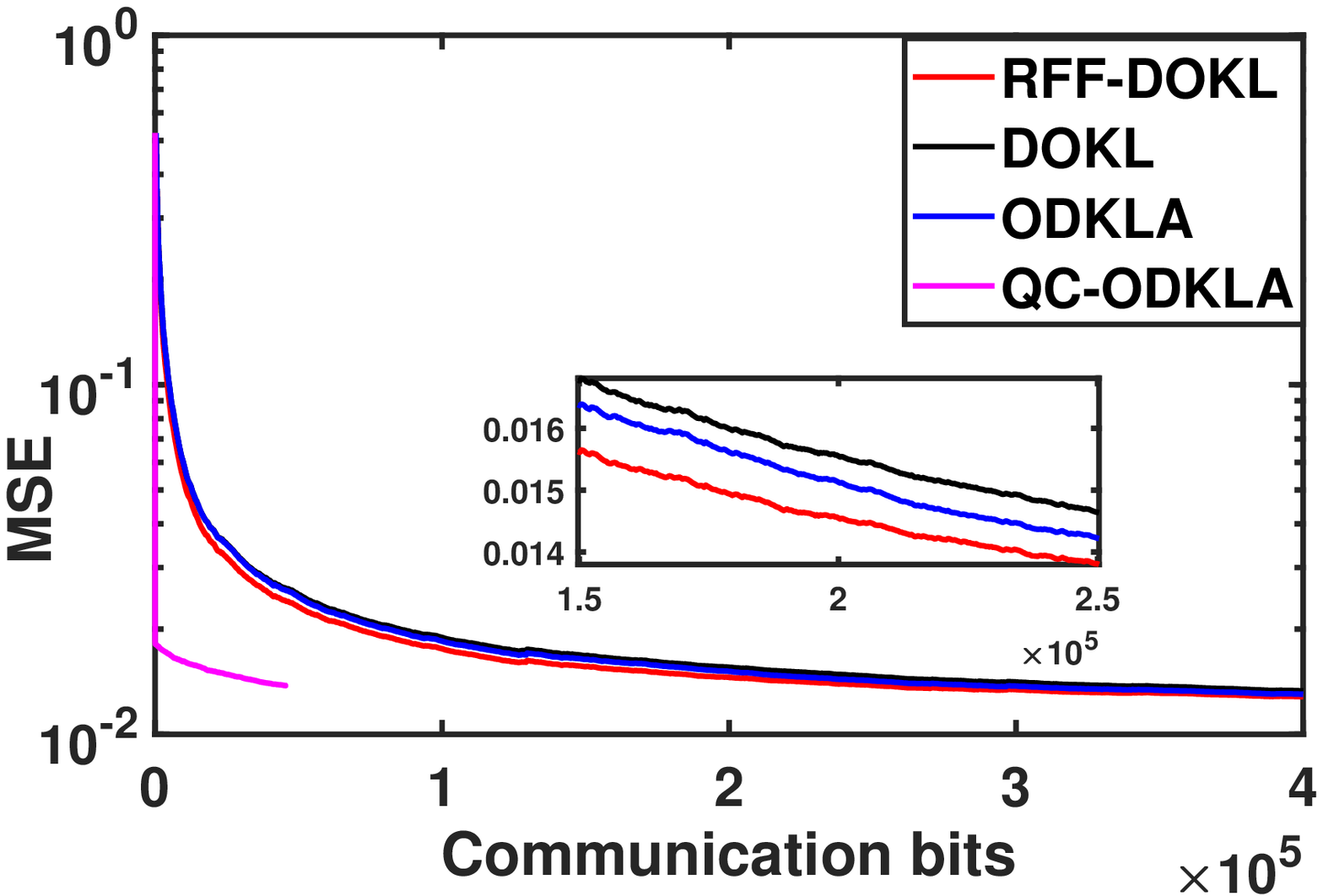}}
	\caption{Learning performance on Conductivity dataset. (a) MSE vs. time; (b) MSE vs. triggers; (c) MSE vs. bits.}\label{fig:conduct}
\end{figure*}

\begin{figure*}[!t]
	\centering
	\subfloat[]{\includegraphics[width=2.2in]{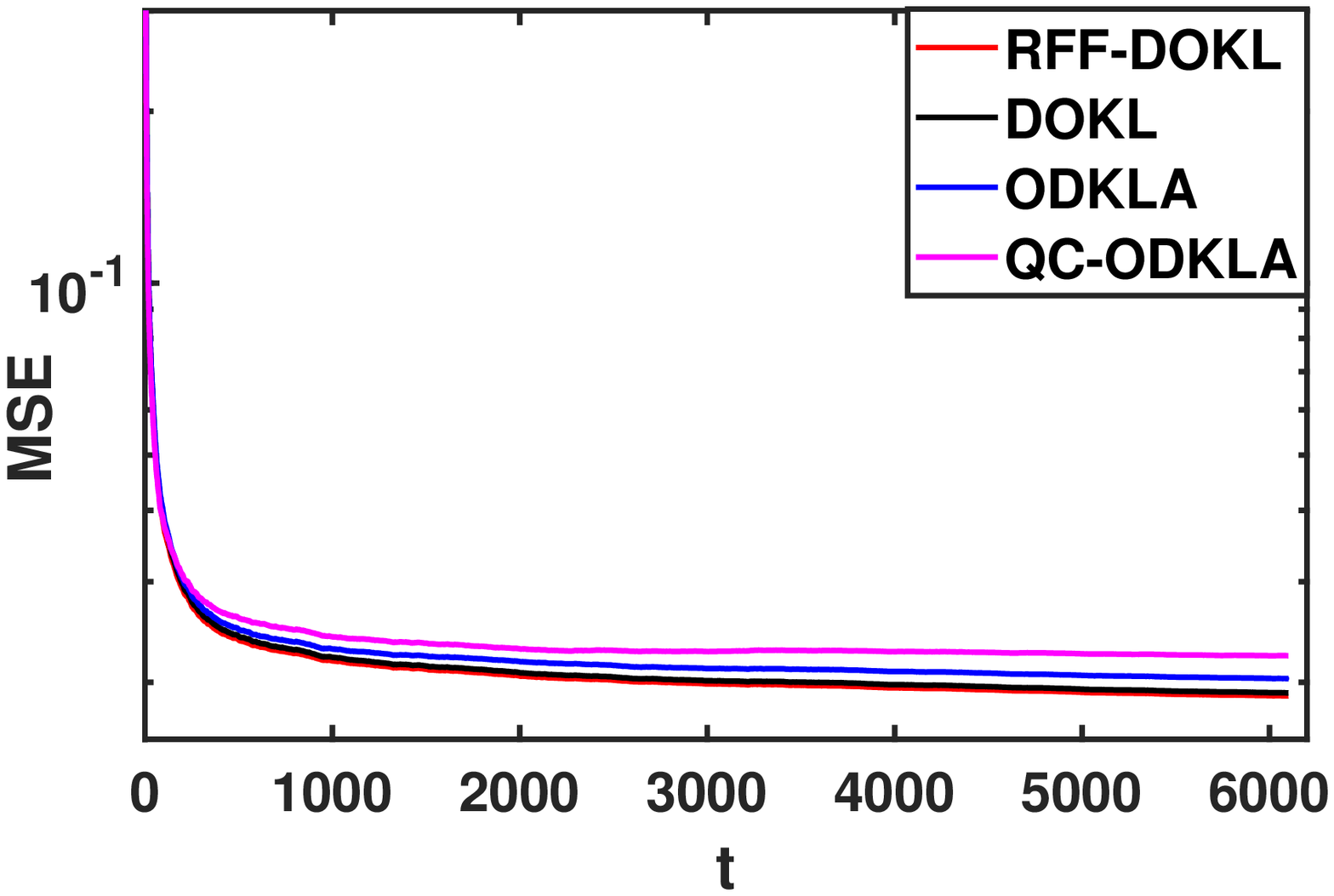}}
	\hfil
	\subfloat[]{\includegraphics[width=2.2in]{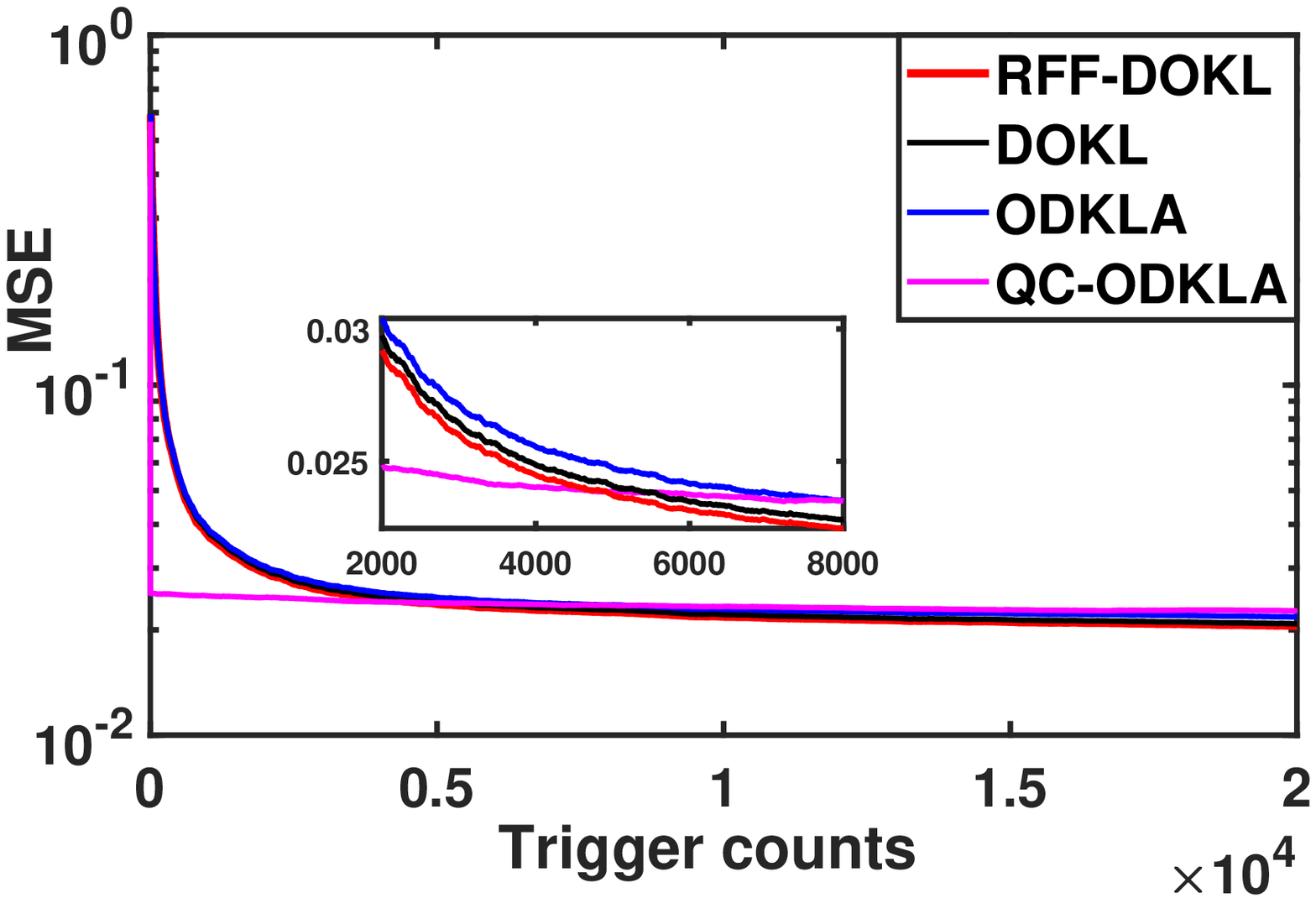}}
	\hfil
	\subfloat[]{\includegraphics[width=2.2in]{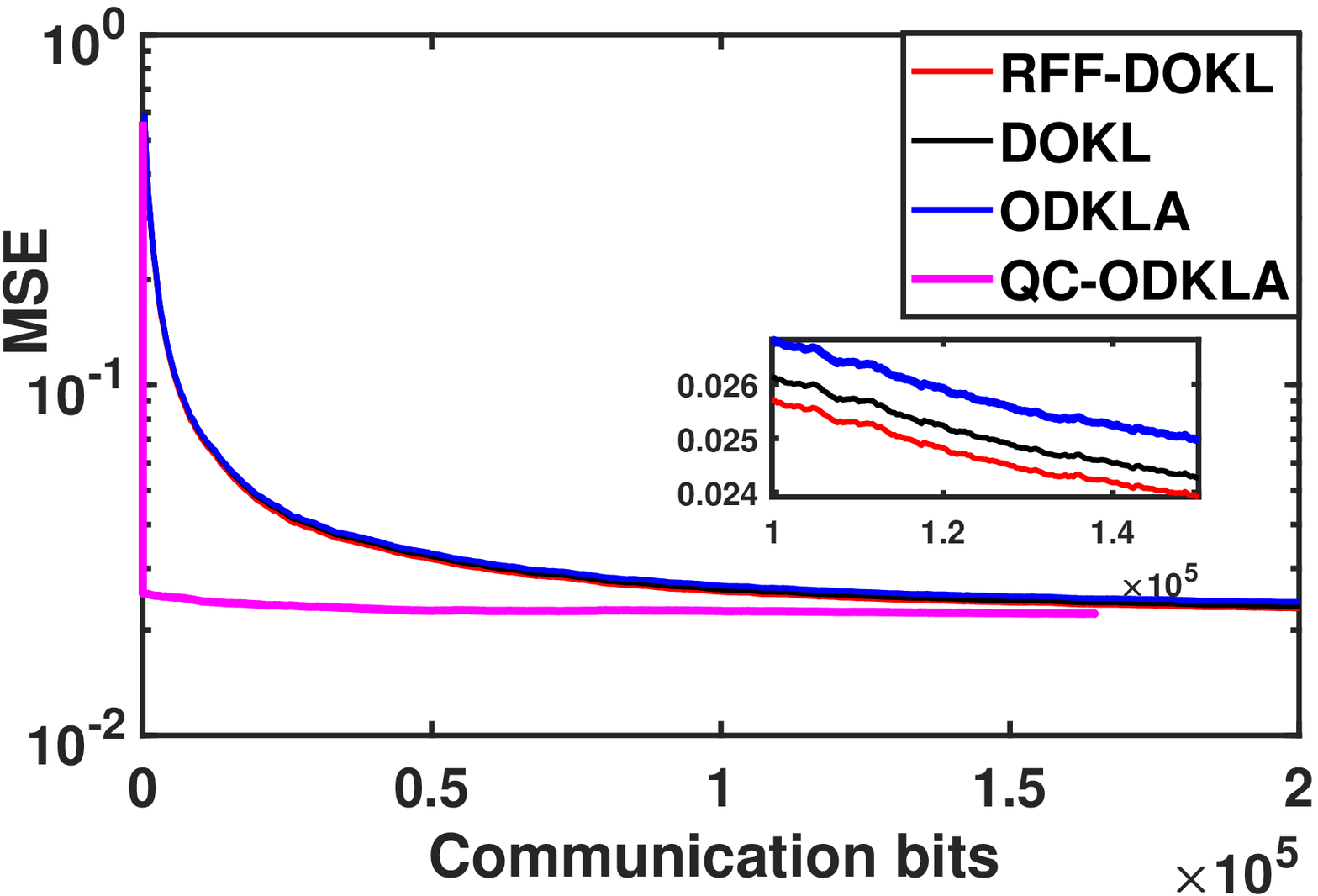}}
	\caption{Learning performance on Blood dataset. (a) MSE vs. time; (b) MSE vs. triggers; (c) MSE vs. bits.}\label{fig:blood}
\end{figure*}

\textbf{MSE performance.} We first evaluate the learning performance of all algorithms by the mean-squared-error (MSE), which is commonly adopted in online learning problems~\cite{bouboulis2018online, hong2021distributed}. From Figures \ref{fig:tom} (a) - \ref{fig:blood} (a), we can see that the learning performance of ODKLA, RFF-DOKL, and DOKL is very close while the trivial difference comes from the distinction of specific datasets. Further, the learning performance of QC-ODKLA is always comparable to that of the ODKLA, after introducing the communication censoring and quantization strategies. The quantization level is set to be $q = 8$,

\textbf{Communication efficiency.} We then evaluate the communication efficiency among different algorithms. We present the MSE performance versus trigger counts in Figures \ref{fig:tom} (b) - \ref{fig:blood} (b) and MSE performance versus communication bits in Figures \ref{fig:tom} (c) - \ref{fig:blood} (c). Figures \ref{fig:tom} (b) - \ref{fig:blood} (b) show that QC-ODKLA triggers a few transmissions in the early learning stage, which greatly improves the communication efficiency. Further, thanks to the quantization, QC-ODKLA only needs 3 bits to transmit an element, the total number of communication bits is also greatly reduced accordingly. For other methods to transmit each element of updates, suppose the agent uses a 32-bit CPU operating mode, then the communication cost is 32 bits per iteration per agent per element. Therefore, QC-ODKLA is corroborated to greatly reduce the communication cost.

\textbf{Computation efficiency.} Finally, we evaluate the computation efficiency of all algorithms by their running time on six datasets, which is recorded in Table \ref{tab:TwittL}. RFF-DOKL is a gradient descent-based first-order algorithm, which achieves the highest computation efficiency. Comparing ODKLA with the ADMM based DOKL method, we see that the linearization step reduces a large amount of computation of standard ADMM. Under the circumstance that online streaming data vary fast, a computation-efficient algorithm is preferred, reflecting the advantages of the proposed ODKLA and QC-ODKLA algorithms. Also, note that QC-ODKLA is computationally slower than ODKLA since the communication censoring and quantization steps consume computation resources.   
 
\setlength{\tabcolsep}{0.5mm}{
	\begin{table}
		\centering
		\caption{The running time of four algorithms on six datasets.	\label{tab:TwittL}} 	
		\begin{tabular}{lcccr}
			\hline
			Data set & RFF-DOKL & DOKL & ODKLA & QC-ODKLA \\
			\hline
			Tom's     & 0.3541s & 2.0613s   & 0.4455s  & 0.7247s \\
			Twitter           &4.6148s & 21.8808s  & 6.2351s  & 9.4218s \\
			Energy            &0.8584s & 4.2681s   & 1.1600s  & 1.7928s \\
			Air quality       &0.2760s & 1.6808s   & 0.4717s  & 0.5510s  \\
			Conductivity      &0.7952s & 4.4285s   & 0.9784s   & 1.6201s   \\
			Blood             &2.6194s & 13.6249s   &3.6104s  & 5.4290s\\
			\hline
		\end{tabular}	
\end{table}}

\section{Conclusion}
\label{sec:con}
This paper studies the online decentralized kernel learning problem under communication constraints for multi-agent systems. We utilize RF mapping to circumvent the curse of dimensionality issue caused by the increasing size of sequentially arriving data. To efficiently solve such a challenging problem, we then develop a novel online decentralized kernel learning algorithm via linearized ADMM (ODKLA). We integrate the communication-censoring and quantization strategies into the proposed ODKAL algorithm (QC-ODKLA) to further save communication overheads. We derive the sublinear regret bound for QC-ODKLA theoretically, and verify their effectiveness in learning performance, communication and computation efficiency via simulations on various real datasets. Future work will be devoted to multi-kernel learning and dynamic kernel learning.

%\section*{Acknowledgments}
%This should be a simple paragraph before the References to thank those individuals and institutions who have supported your work on this article.

{\appendix
\section{Proof of Theorem \ref{theor:theorm2} }
\label{subsec:theorem2}
\textit{Proof.} Define $\bm{\Theta}^\star = [\bm{\theta}^{\star\top};\dots;\bm{\theta}^{\star\top}]\in \mathbb{R}^{N\times 2L}$, which is the stack of $N$ copies of $\bm{\theta}^{\star}$, and $\mathcal{L}_{t}(\bm{\Theta}^\star):= \sum_{i=1}^N \mathcal{L}_{i,t}(\bm{\theta}^\star)$, we rewrite \eqref{eq:stat_reg} as
\begin{equation}
	\label{eq:static_reg_mtrx}
	\begin{split}
		\mathbf{Reg}_T^S  &  = \sum_{t=1}^T \left(\sum_{i=1}^N \mathcal{L}_{i,t}(\bm{\theta}_{i,t}) - \sum_{i=1}^N \mathcal{L}_{i,t}(\bm{\theta}^\star) \right) \\
		& =  \sum_{t=1}^T \left( \mathcal{L}_{t}(\bm{\Theta}_{t}) - \mathcal{L}_{t}(\bm{\Theta}^\star) \right). 
	\end{split}
\end{equation} 

To analyze the regret bound of QC-ODKLA, we first represent the matrix form update of QC-ODKLA updates \eqref{eq:dec_qcodkla_primal} - \eqref{eq:dec_qcodkla_dual} as
\begin{align}
	\begin{split}
		\label{eq:qcoke_primal}
		\bm{\Theta}_{t+1} &= \bm{\Theta}_{t} -  (\eta_t\mathbf{I} + 2\rho\bm{D})^{-1}\Big[ \partial \mathcal{L}_{t} (\bm{\Theta}_{t})  \\
		 & \qquad \qquad \qquad \qquad \qquad +  \rho (\bm{D}-\bm{W}) \hat{\bm{\Theta}}_{t}   +\bm{\Gamma}_{t}  \Big],
	\end{split}\\
	\begin{split}
		\label{eq:qcoke_dual}
		\bm{\Gamma}_{t+1} &= \bm{\Gamma}_{t} + \rho (\bm{D}-\bm{W}) \hat{\bm{\Theta}}_{t+1},
	\end{split}
\end{align}
where $\hat{\bm{\Theta}}_t = [\hat{\bm{\theta}}_{1,t}^\top;\dots;\hat{\bm{\theta}}_{N,t}^\top]\in \mathbb{R}^{N\times 2L}$. Note that the censoring and quantization are implemented after step \eqref{eq:qcoke_primal} and before step \eqref{eq:qcoke_dual}. 

The definitions of the introduced error in \eqref{eq:intro_error_i} and the overall introduced error $\bm{E}_t$ is equivalent to $\bm{E}_t: = \bm{\Theta}_t - \hat{\bm{\Theta}}_t$. With the equality $\mathbf{D}-\mathbf{W} = \frac{1}{2}\mathbf{S}_{-}\mathbf{S}_{-}^\top$, we can obtain the equivalent form of \eqref{eq:qcoke_primal} and \eqref{eq:qcoke_dual} respectively as
\begin{align}
	\begin{split}
		\label{eq:equiv_qcoke_primal}
		\bm{\Theta}_{t+1} &= \bm{\Theta}_{t} -  (\eta_t\mathbf{I} + 2\rho\bm{D})^{-1}\\
		& \quad \Big[\partial \mathcal{L}_{t} (\bm{\Theta}_{t}) + \frac{\rho}{2}\mathbf{S}_{-}\mathbf{S}_{-}^\top\bm{\Theta}_t -  \frac{\rho}{2}\mathbf{S}_{-}\mathbf{S}_{-}^\top \bm{E}_t +\bm{\Gamma}_{t}  \Big],
	\end{split}\\
	\begin{split}
		\label{eq:equiv_qcoke_dual}
		\bm{\Gamma}_{t+1} &= \bm{\Gamma}_{t} + \frac{\rho}{2}\mathbf{S}_{-}\mathbf{S}_{-}^\top\bm{\Theta}_{t+1} - \frac{\rho}{2}\mathbf{S}_{-}\mathbf{S}_{-}^\top \bm{E}_{t+1}.
	\end{split}
\end{align}
Observe from \eqref{eq:equiv_qcoke_dual} that $\bm{\Gamma}_{t+1}$ stays in the column space of $\mathbf{S}_{-}\mathbf{S}_{-}^\top$ if $\bm{\Gamma}_{1}$ is also initialized therein. Therefore, we introduce variables $\bm{\beta}_t\in \mathbb{R}^{2r\times 2L}$, which stay in the column space of $\mathbf{S}_{-}^\top$, and let $\bm{\Gamma}_{t}  = \mathbf{S}_{-}\bm{\beta}_t$ for any $t\geq 1$. Then, \eqref{eq:equiv_qcoke_dual} is equivalent to 
\begin{equation}
	\label{eq:equiv_qcoke_dual_2}
	\bm{\beta}_{t+1} = \bm{\beta}_{t} +\frac{\rho}{2}\mathbf{S}^\top_{-}\bm{\Theta}_{t+1} - \frac{\rho}{2}\mathbf{S}_{-}^\top \bm{E}_{t+1}. 
\end{equation}
Using \eqref{eq:equiv_qcoke_dual} and $\bm{\Gamma}_{t}  = \mathbf{S}_{-}\bm{\beta}_{t}$ to eliminate $\bm{\Gamma}_{t}$, we rewrite \eqref{eq:equiv_qcoke_primal} as
\begin{equation}
	\label{eq:equiv_qcoke_primal_2} 
	\begin{split}
	&\bm{\Theta}_{t+1} = \bm{\Theta}_{t} -  (\eta_t\mathbf{I} + 2\rho\bm{D})^{-1}  \Big[ \partial \mathcal{L}_{t} (\bm{\Theta}_{t})  + \mathbf{S}_{-}\bm{\beta}_{t+1}  \\
	& \quad \; \;+ \frac{\rho}{2}\mathbf{S}_{-}\mathbf{S}_{-}^\top(\bm{E}_{t+1}  -\bm{E}_{t})  + \frac{\rho}{2}\mathbf{S}_{-}\mathbf{S}_{-}^\top (\bm{\Theta}_t -\bm{\Theta}_{t+1}) \Big].
	\end{split}
\end{equation}

The following analysis is based on the equivalent form of the QC-ODKLA algorithm given by \eqref{eq:equiv_qcoke_primal_2} and \eqref{eq:equiv_qcoke_dual_2}. The Karush–Kuhn–Tucker (KKT) conditions of \eqref{eq:online_opt_matrix} are
\begin{subequations}
	\begin{align}
		\partial  \mathcal{L}_{t}(\bm{\Theta}^\star) + \eta_t(\bm{\Theta}^\star - \bm{\Theta}_t) + \mathbf{S}_{-}\bm{\beta}^\star & = \mathbf{0}, \label{eq:kkt_1}\\
		\mathbf{S}_{-}^\top \bm{\Theta}^\star & = \mathbf{0}, \label{eq:kkt_2}\\
		\frac{1}{2}\mathbf{S}_{+}^\top\bm{\Theta}^\star & =  \bm{Z}^\star, \label{eq:kkt_3}
	\end{align}
\end{subequations}
where $(\bm{\Theta}^\star, \bm{Z}^\star, \bm{\beta}^\star)$ is the optimal primal-dual triplet. 

Rearrange terms in \eqref{eq:equiv_qcoke_primal_2} to place $\partial \mathcal{L}_{t} (\bm{\Theta}_{t})$ at the left side, we have
\begin{equation}
	\label{eq:partial_theta}
	\begin{split}
		\partial \mathcal{L}_{t} (\bm{\Theta}_{t}) 
		&= (\eta_t\mathbf{I} + 2\rho\bm{D} - \frac{\rho}{2}\mathbf{S}_{-}\mathbf{S}_{-}^\top) (\bm{\Theta}_t -\bm{\Theta}_{t+1}) \\
		& \quad + \frac{\rho}{2}\mathbf{S}_{-}\mathbf{S}_{-}^\top(\bm{E}_{t}  -\bm{E}_{t+1}) - \mathbf{S}_{-}\bm{\beta}_{t+1}\\
		&=  (\eta_t\mathbf{I} + \frac{\rho}{2}\mathbf{S}_{+}\mathbf{S}_{+}^\top) (\bm{\Theta}_t -\bm{\Theta}_{t+1}) \\
		& \quad + \frac{\rho}{2}\mathbf{S}_{-}\mathbf{S}_{-}^\top(\bm{E}_{t}  -\bm{E}_{t+1}) - \mathbf{S}_{-}\bm{\beta}_{t+1}, 
	\end{split}
\end{equation}
where the second equality utilizes $\mathbf{D}-\mathbf{W} = \frac{1}{2}\mathbf{S}_{-}\mathbf{S}_{-}^\top$ and $\mathbf{D}+\mathbf{W} = \frac{1}{2}\mathbf{S}_{+}\mathbf{S}_{+}^\top$ such that $2\mathbf{D} =\frac{1}{2}\mathbf{S}_{-}\mathbf{S}_{-}^\top + \frac{1}{2}\mathbf{S}_{+}\mathbf{S}_{+}^\top$.
We consider to bound the instantaneous regret $\mathcal{L}_{t}(\bm{\Theta}_{t}) - \mathcal{L}_{t}(\bm{\Theta}^\star)$ at time $t$ first. With Assumption \ref{ass:convex}, it holds 
\begin{equation}
	\label{eq:stat_ins}
	\mathcal{L}_{t}(\bm{\Theta}_{t})- \mathcal{L}_{t}(\bm{\Theta}^\star)\leq  \langle   \partial\mathcal{L}_{t}(\bm{\Theta}_{t}), \bm{\Theta}_{t} - \bm{\Theta}^\star    \rangle.
\end{equation} 
 
%We consider to bound the instantaneous regret $\mathcal{L}_{t}(\bm{\Theta}_{t}) - \mathcal{L}_{t}(\bm{\Theta}^\star)$ at time $t$ first. With assumption 1, it holds 
%\begin{equation}
%	\label{eq:stat_ins}
%	\mathcal{L}_{t}(\bm{\Theta}_{t})- \mathcal{L}_{t}(\bm{\Theta}^\star)\leq  \langle   \partial\mathcal{L}_{t}(\bm{\Theta}_{t}), \bm{\Theta}_{t} - \bm{\Theta}^\star    \rangle.
%\end{equation} 
%
Substitute the expression of  $ \partial\mathcal{L}_{t}(\bm{\Theta}_{t})$ in \eqref{eq:partial_theta} into \eqref{eq:stat_ins} yields
\begin{equation}
	\label{eq:stat_ins_1}
	\begin{split}
		&\quad \mathcal{L}_{t}(\bm{\Theta}_{t})- \mathcal{L}_{t}(\bm{\Theta}^\star) \\
		&\leq \langle (\eta_t\mathbf{I} +  \frac{\rho}{2}\mathbf{S}_{+}\mathbf{S}_{+}^\top) (\bm{\Theta}_t -\bm{\Theta}_{t+1}), \bm{\Theta}_{t} - \bm{\Theta}^\star  \rangle   \\
		&+ \langle \frac{\rho}{2} \mathbf{S}_{-}\mathbf{S}_{-}^\top(\bm{E}_{t} -\bm{E}_{t+1})-\mathbf{S}_{-}\bm{\beta}_{t+1}, \bm{\Theta}_{t} - \bm{\Theta}^\star    \rangle.   
	\end{split}
\end{equation}
Now we reorganize the two terms on the right-hand side of \eqref{eq:stat_ins_1}. For the first term, we have
\begin{equation}
	\label{eq:stat_ins_2}
	\begin{split}
	& \quad 	\langle (\eta_t\mathbf{I} +  \frac{\rho}{2}\mathbf{S}_{+}\mathbf{S}_{+}^\top) (\bm{\Theta}_t -\bm{\Theta}_{t+1}), \bm{\Theta}_{t} - \bm{\Theta}^\star  \rangle \\
		&\leq \sigma_{\max}(\eta_t\mathbf{I} +  \frac{\rho}{2}\mathbf{S}_{+}\mathbf{S}_{+}^\top) \langle\bm{\Theta}_t -\bm{\Theta}_{t+1},    \bm{\Theta}_{t} - \bm{\Theta}^\star    \rangle\\
		&=\frac{\sigma_{\max}(\eta_t\mathbf{I} +  \frac{\rho}{2}\mathbf{S}_{+}\mathbf{S}_{+}^\top)}{2} \Big(\| \bm{\Theta}_{t} -\bm{\Theta}^\star\|_F^2  \\
		& \quad - \|\bm{\Theta}_{t+1} - \bm{\Theta}^\star \|_F^2 + \|\bm{\Theta}_{t}- \bm{\Theta}_{t+1} \|_F^2\Big),
	\end{split}
\end{equation}
where $\sigma_{\max}(\eta_t\mathbf{I} +  \frac{\rho}{2}\mathbf{S}_{+}\mathbf{S}_{+}^\top)$ denotes the maximum singular value of $\eta_t\mathbf{I} +  \frac{\rho}{2}\mathbf{S}_{+}\mathbf{S}_{+}^\top$.

For the second term, we have
\begin{equation}
	\label{eq:stat_ins_3}
	\begin{split}
		&\langle \frac{\rho}{2} \mathbf{S}_{-}\mathbf{S}_{-}^\top(\bm{E}_{t} -\bm{E}_{t+1})-\mathbf{S}_{-}\bm{\beta}_{t+1}, \bm{\Theta}_{t} -  \bm{\Theta}^\star    \rangle \\
		& =   \langle \frac{\rho}{2}  \mathbf{S}_{-}^\top(\bm{E}_{t} -\bm{E}_{t+1})- \bm{\beta}_{t+1}, \mathbf{S}_{-}^\top(\bm{\Theta}_{t} - \bm{\Theta}^\star)    \rangle \\
		& \stackrel{(a)}{=}  \langle \frac{\rho}{2}  \mathbf{S}_{-}^\top(\bm{E}_{t} -\bm{E}_{t+1})- \bm{\beta}_{t+1}, \mathbf{S}_{-}^\top \bm{\Theta}_{t}   \rangle \\
		& \stackrel{(b)}{=}  \langle \frac{\rho}{2}  \mathbf{S}_{-}^\top(\bm{E}_{t} -\bm{E}_{t+1})- \bm{\beta}_{t+1},  \frac{2}{\rho}( \bm{\beta}_{t} -\bm{\beta}_{t-1}) + \mathbf{S}_{-}^\top \bm{E}_{t} \rangle\\
		& = \langle \bm{\beta}_{t-1} - 2\bm{\beta}_{t} + \frac{\rho}{2}  \mathbf{S}_{-}^\top (\bm{\Theta}_t -\bm{\Theta}_{t+1}) ,  \frac{2}{\rho}( \bm{\beta}_{t} -\bm{\beta}_{t-1}) + \mathbf{S}_{-}^\top \bm{E}_{t} \rangle   \\
		& \stackrel{(c)}{=}  -\frac{2}{\rho} \langle \bm{\beta}_{t} - \bm{\beta}_{t-1}, \bm{\beta}_{t} -\bm{\beta}_{t-1} \rangle - \frac{2}{\rho}\langle\bm{\beta}_{t},  \bm{\beta}_{t} -\bm{\beta}_{t-1} \rangle \\
		&\quad +  \langle \bm{\beta}_{t-1} - \bm{\beta}_{t}, \mathbf{S}_{-}^\top \bm{E}_{t}  \rangle  + \langle   \mathbf{S}_{-}^\top (\bm{\Theta}_t -\bm{\Theta}_{t+1}) , \bm{\beta}_{t} -\bm{\beta}_{t-1}  \rangle   \\
		& \quad + \frac{\rho}{2}  \langle   \mathbf{S}_{-}^\top (\bm{\Theta}_t -\bm{\Theta}_{t+1}) , \mathbf{S}_{-}^\top \bm{E}_{t} \rangle - \langle  \bm{\beta}_{t}, \mathbf{S}_{-}^\top \bm{E}_{t}\rangle 		\\
		&= -\frac{2}{\rho}  \| \bm{\beta}_{t}- \bm{\beta}_{t-1}\|_F^2  - \frac{2}{\rho} \| \bm{\beta}_{t}\|_F^2  +\frac{2}{\rho}  \langle \bm{\beta}_{t}, \bm{\beta}_{t-1}  \rangle  - \langle  \bm{\beta}_{t}, \mathbf{S}_{-}^\top \bm{E}_{t}\rangle  \\
		&\quad + \langle \bm{\beta}_{t-1} - \bm{\beta}_{t}, \mathbf{S}_{-}^\top \bm{E}_{t}  \rangle  + \langle   \mathbf{S}_{-}^\top (\bm{\Theta}_t -\bm{\Theta}_{t+1}) , \bm{\beta}_{t} -\bm{\beta}_{t-1}  \rangle   \\
		& \quad	+ \frac{\rho}{2}  \langle   \mathbf{S}_{-}^\top (\bm{\Theta}_t -\bm{\Theta}_{t+1}) , \mathbf{S}_{-}^\top \bm{E}_{t} \rangle, \\  
	\end{split}
\end{equation}
where (a) comes from the KKT condition \eqref{eq:kkt_2}, (b) and (c) are obtained by utilizing \eqref{eq:equiv_qcoke_dual_2}.

Next, we will utilize Young's inequality to bound the inner product terms in \eqref{eq:stat_ins_3}, which are
\begin{equation}
\label{eq:youngs}
\begin{split}
&\frac{2}{\rho}  \langle \bm{\beta}_{t}, \bm{\beta}_{t-1}  \rangle  \leq \frac{2}{\rho} ( \frac{1}{2\eta_1} \| \bm{\beta}_{t}\|_F^2  +  \frac{\eta_1}{2} \| \bm{\beta}_{t-1}\|_F^2) \\ & \qquad \qquad \quad  = \frac{1}{\rho\eta_1}  \| \bm{\beta}_{t}\|_F^2 +  \frac{\eta_1}{\rho}  \| \bm{\beta}_{t-1}\|_F^2, \\
&\langle \bm{\beta}_{t-1} - \bm{\beta}_{t}, \mathbf{S}_{-}^\top \bm{E}_{t}\rangle \leq  \frac{1}{2\eta_2} \|\bm{\beta}_{t-1} - \bm{\beta}_{t}\|_F^2  + \frac{\eta_2}{2} \|\mathbf{S}_{-}^\top \bm{E}_{t}\|_F^2 , \\%
&-\langle  \bm{\beta}_{t}, \mathbf{S}_{-}^\top \bm{E}_{t}\rangle  \leq \frac{1}{2\eta_3} \|\bm{\beta}_{t}\|_F^2  + \frac{\eta_3}{2} \|\mathbf{S}_{-}^\top \bm{E}_{t}\|_F^2, \\
&\langle   \mathbf{S}_{-}^\top (\bm{\Theta}_t -\bm{\Theta}_{t+1}) , \bm{\beta}_{t} -\bm{\beta}_{t-1}  \rangle  \leq \frac{1}{2\eta_4} \|\mathbf{S}_{-}^\top (\bm{\Theta}_t -\bm{\Theta}_{t+1})\|_F^2  \\& \qquad \qquad \quad \qquad \qquad \quad  \qquad  \quad   + \frac{\eta_4}{2} \|\bm{\beta}_{t} -\bm{\beta}_{t-1} \|_F^2, \\
&\frac{\rho}{2}  \langle   \mathbf{S}_{-}^\top (\bm{\Theta}_t -\bm{\Theta}_{t+1}) , \mathbf{S}_{-}^\top \bm{E}_{t} \rangle   \leq  \frac{\rho}{4\eta_5}\|\mathbf{S}_{-}^\top (\bm{\Theta}_t -\bm{\Theta}_{t+1})\|_F^2 \\ & \qquad \qquad \qquad \qquad \qquad \qquad + \frac{\rho\eta_5}{4}\|\mathbf{S}_{-}^\top \bm{E}_{t} \|_F^2,		
\end{split}
\end{equation}
where $\eta_1, \eta_2,\eta_3,\eta_4,\eta_5$ are any positive constants. 

Substitute \eqref{eq:youngs} into \eqref{eq:stat_ins_3} gives
\begin{equation}
	\label{eq:stat_ins_4}
	\begin{split}
		&\langle \frac{\rho}{2} \mathbf{S}_{-}\mathbf{S}_{-}^\top(\bm{E}_{t} -\bm{E}_{t+1})-\mathbf{S}_{-}\bm{\beta}_{t+1}, \bm{\Theta}_{t} - \bm{\Theta}^\star \rangle \\
		&\leq (-\frac{2}{\rho} + \frac{1}{2\eta_2} + \frac{\eta_4}{2} ) \| \bm{\beta}_{t}- \bm{\beta}_{t-1}\|_F^2 + \frac{\eta_1}{\rho}  \| \bm{\beta}_{t-1}\|_F^2\\
		        & \quad + (- \frac{2}{\rho} +  \frac{1}{\rho\eta_1} + \frac{1}{2\eta_3}) \| \bm{\beta}_{t}\|_F^2  \\
	          	& \quad + ( \frac{1}{2\eta_4} +  \frac{\rho}{4\eta_5})\|\mathbf{S}_{-}^\top (\bm{\Theta}_t -\bm{\Theta}_{t+1})\|_F^2 \\
	          	& \quad + ( \frac{\eta_2}{2} + \frac{\eta_3}{2} + \frac{\rho\eta_5}{4})\|\mathbf{S}_{-}^\top \bm{E}_{t} \|_F^2 \\
		& = (-\frac{2}{\rho} + \frac{1}{2\eta_2} + \frac{\eta_4}{2} ) ( \| \bm{\beta}_{t}\|_F^2  + \| \bm{\beta}_{t-1}\|_F^2 - 2\langle \bm{\beta}_{t},\bm{\beta}_{t-1} \rangle) \\
	        	&\quad + (- \frac{2}{\rho} +  \frac{1}{\rho\eta_1} + \frac{1}{2\eta_3}) \| \bm{\beta}_{t}\|_F^2 + \frac{\eta_1}{\rho}  \| \bm{\beta}_{t-1}\|_F^2 \\
		        &\quad + ( \frac{1}{2\eta_4} +  \frac{\rho}{4\eta_5})\|\mathbf{S}_{-}^\top (\bm{\Theta}_t -\bm{\Theta}_{t+1})\|_F^2 \\
				& \quad + ( \frac{\eta_2}{2} + \frac{\eta_3}{2} + \frac{\rho\eta_5}{4})\|\mathbf{S}_{-}^\top \bm{E}_{t} \|_F^2\\
		& = ( -\frac{4}{\rho} + \frac{1}{2\eta_2} + \frac{\eta_4}{2} +  \frac{1}{\rho\eta_1} + \frac{1}{2\eta_3})\| \bm{\beta}_{t}\|_F^2 \\
			&\quad +  (-\frac{2}{\rho} + \frac{1}{2\eta_2} + \frac{\eta_4}{2} +\frac{\eta_1}{\rho}  )\| \bm{\beta}_{t-1}\|_F^2  \\
			&\quad + ( \frac{1}{2\eta_4} +  \frac{\rho}{4\eta_5})\|\mathbf{S}_{-}^\top (\bm{\Theta}_t -\bm{\Theta}_{t+1})\|_F^2 \\
		    &\quad + ( \frac{\eta_2}{2} + \frac{\eta_3}{2} + \frac{\rho\eta_5}{4})\|\mathbf{S}_{-}^\top \bm{E}_{t} \|_F^2   \\
		    &\quad +  (\frac{2}{\rho} - \frac{1}{2\eta_2} - \frac{\eta_4}{2} )\langle \bm{\beta}_{t},\bm{\beta}_{t-1} \rangle \\
		&\leq \Big( -\frac{4}{\rho} + \frac{1}{2\eta_2} + \frac{\eta_4}{2} +  \frac{1}{\rho\eta_1} + \frac{1}{2\eta_3} + \frac{2}{\rho \eta_6} \\
	    	& \qquad- \frac{1}{2\eta_2\eta_6} - \frac{\eta_4}{2\eta_6} \Big)\| \bm{\beta}_{t}\|_F^2 \\
		    &\quad +  \Big(-\frac{2}{\rho} + \frac{1}{2\eta_2} + \frac{\eta_4}{2} +\frac{\eta_1}{\rho} + \frac{2\eta_6}{\rho } \\
		    & \qquad - \frac{\eta_6}{2\eta_2} - \frac{\eta_4\eta_6}{2}  \Big) \| \bm{\beta}_{t-1}\|_F^2  \\
		    &\quad + \Big( \frac{1}{2\eta_4} +  \frac{\rho}{4\eta_5}\Big)\|\mathbf{S}_{-}^\top (\bm{\Theta}_t -\bm{\Theta}_{t+1})\|_F^2 \\
		    & \quad + \Big( \frac{\eta_2}{2} + \frac{\eta_3}{2} + \frac{\rho\eta_5}{4}\Big)\|\mathbf{S}_{-}^\top \bm{E}_{t} \|_F^2. \\
		%& = c(\| \bm{\beta}_{t-1}\|_F^2  - \| \bm{\beta}_{t}\|_F^2 ) + ( \frac{1}{2\eta_4} +  \frac{\rho}{4\eta_5})\|\mathbf{S}_{-}^\top (\bm{\Theta}_t -\bm{\Theta}_{t+1})\|_F^2 + ( \frac{\eta_2}{2} + \frac{\eta_3}{2} + \frac{\rho\eta_5}{4})\|\mathbf{S}_{-}^\top \bm{E}_{t} \|_F^2,
	\end{split}
\end{equation}
%where the last equality can be obtained by setting 
%\begin{equation}
%- ( -\frac{4}{\rho} + \frac{1}{2\eta_2} + \frac{\eta_4}{2} +  \frac{1}{\rho\eta_1} + \frac{1}{2\eta_3} + \frac{2}{\rho \eta_6} - \frac{1}{2\eta_2\eta_6} - \frac{\eta_4}{2\eta_6} ) =  (-\frac{2}{\rho} + \frac{1}{2\eta_2} + \frac{\eta_4}{2} +\frac{\eta_1}{\rho} + \frac{2\eta_6}{\rho } - \frac{\eta_6}{2\eta_2} - \frac{\eta_4\eta_6}{2}  ) =c.
%\end{equation} 
%The above equality can be easily achieved. For example, for $\rho =1/2$, we can find $eta_1 = \eta_2 =\eta_3=1$, $\eta_4 = 9$, and $\eta_6 = 1/2$.  

With \eqref{eq:stat_ins_4} and \eqref{eq:stat_ins_2}, we obtain an upper bound for \eqref{eq:stat_ins_1}, which is
\begin{equation}
	\label{eq:stat_ins_com}
	\begin{split}
		&\mathcal{L}_{t}(\bm{\Theta}_{t})- \mathcal{L}_{t}(\bm{\Theta}^\star)\\
		&\leq \frac{\sigma_{\max}(\eta_t\mathbf{I} +  \frac{\rho}{2}\mathbf{S}_{+}\mathbf{S}_{+}^\top)}{2} \Big(\|\bm{\Theta}_{t} -\bm{\Theta}^\star\|_F^2 - \|\bm{\Theta}_{t+1} - \bm{\Theta}^\star \|_F^2\Big)  \\
	     	&\quad +  \frac{\sigma_{\max}(\eta_t\mathbf{I} +  \frac{\rho}{2}\mathbf{S}_{+}\mathbf{S}_{+}^\top)}{2} \|\bm{\Theta}_{t}- \bm{\Theta}_{t+1} \|_F^2 \\
		    &\quad  + (\frac{1}{2\eta_2}-\frac{4}{\rho} + \frac{\eta_4}{2} +  \frac{1}{\rho\eta_1} + \frac{1}{2\eta_3} + \frac{2}{\rho \eta_6} - \frac{1}{2\eta_2\eta_6} - \frac{\eta_4}{2\eta_6} )\| \bm{\beta}_{t}\|_F^2\\
		    &\quad +  (\frac{\eta_1}{\rho}-\frac{2}{\rho} + \frac{1}{2\eta_2} + \frac{\eta_4}{2} + \frac{2\eta_6}{\rho } - \frac{\eta_6}{2\eta_2} - \frac{\eta_4\eta_6}{2}) \| \bm{\beta}_{t-1}\|_F^2  \\
	     	& \quad  + ( \frac{1}{2\eta_4} + \frac{\rho}{4\eta_5})\|\mathbf{S}_{-}^\top (\bm{\Theta}_t -\bm{\Theta}_{t+1})\|_F^2 + ( \frac{\eta_2}{2} + \frac{\eta_3}{2} + \frac{\rho\eta_5}{4})\|\mathbf{S}_{-}^\top \bm{E}_{t} \|_F^2 \\
		&\leq \frac{\sigma_{\max}(\eta_t\mathbf{I} +  \frac{\rho}{2}\mathbf{S}_{+}\mathbf{S}_{+}^\top)}{2} \Big(\|\bm{\Theta}_{t} -\bm{\Theta}^\star\|_F^2 - \|\bm{\Theta}_{t+1} - \bm{\Theta}^\star \|_F^2\Big) \\
	    	&\quad + ( \frac{1}{2\eta_2} -\frac{4}{\rho} + \frac{\eta_4}{2} +  \frac{1}{\rho\eta_1} + \frac{1}{2\eta_3} + \frac{2}{\rho \eta_6} - \frac{1}{2\eta_2\eta_6} - \frac{\eta_4}{2\eta_6} )\| \bm{\beta}_{t}\|_F^2  \\
		    &\quad  +  (\frac{\eta_1}{\rho}-\frac{2}{\rho} + \frac{1}{2\eta_2} + \frac{\eta_4}{2} + \frac{2\eta_6}{\rho } - \frac{\eta_6}{2\eta_2} - \frac{\eta_4\eta_6}{2}) \| \bm{\beta}_{t-1}\|_F^2 \\
	     	&\quad + (\frac{\sigma_{\max}(\eta_t\mathbf{I} +  \frac{\rho}{2}\mathbf{S}_{+}\mathbf{S}_{+}^\top)}{2}+  \frac{\sigma^2_{\max}(\mathbf{S}_{-})}{2\eta_4} + \frac{\rho\sigma^2_{\max}(\mathbf{S}_{-})}{4\eta_5}) \|\bm{\Theta}_t -\bm{\Theta}_{t+1}\|_F^2 \\
		   &\quad +( \frac{\eta_2}{2} + \frac{\eta_3}{2} +  \frac{\rho\eta_5}{4})\sigma^2_{\max}(\mathbf{S}_{-}) \|\bm{E}_{t} \|_F^2.
	\end{split}
\end{equation}

We then utilize \eqref{eq:equiv_qcoke_primal_2} to rewrite $\bm{\Theta}_t -\bm{\Theta}_{t+1}$ as
\begin{equation}
	\label{eq:theta_rewrite}
	\bm{\Theta}_t -\bm{\Theta}_{t+1} = (\eta_t\mathbf{I} + 2\rho\bm{D})^{-1}\Big( \partial \mathcal{L}_{t} (\bm{\Theta}_{t}) + 2\mathbf{S}_{-}\bm{\beta}_{t} -  \mathbf{S}_{-}\bm{\beta}_{t-1} \Big),
\end{equation}
and bound $\|\bm{\Theta}_t -\bm{\Theta}_{t+1}\|_F^2$ as
\begin{equation}
	\label{eq:theta_norm}
	\begin{split}
		\|\bm{\Theta}_t -\bm{\Theta}_{t+1}\|_F^2 &=\| (\eta_t\mathbf{I} + 2\rho\bm{D})^{-1}\Big( \partial \mathcal{L}_{t} (\bm{\Theta}_{t}) + 2\mathbf{S}_{-}\bm{\beta}_{t} -  \mathbf{S}_{-}\bm{\beta}_{t-1} \Big)\|_F^2 \\
		&\leq \frac{1}{\sigma^2_{\min}(\eta_t\mathbf{I} + 2\rho\bm{D}  ) }\|\partial \mathcal{L}_{t} (\bm{\Theta}_{t})\|_F^2 \\
		& +  \frac{4\sigma^2_{\max}(\mathbf{S}_{-})  }{\sigma^2_{\min}(\eta_t\mathbf{I} + 2\rho\bm{D})}\|\bm{\beta}_{t}\|_F^2 \\
		&+  \frac{\sigma^2_{\max}(\mathbf{S}_{-})}{\sigma^2_{\min}(\eta_t\mathbf{I} + 2\rho\bm{D})} \|\bm{\beta}_{t-1}\|_F^2,
	\end{split}
\end{equation}
where $\sigma_{\min}(\eta_t\mathbf{I} + 2\rho\bm{D})$ is the lower bound of the nonzero singular values of $\eta_t\mathbf{I} + 2\rho\bm{D}$.

Substitute \eqref{eq:theta_norm} into \eqref{eq:stat_ins_com} we obtain
\begin{equation}
	\label{eq:stat_ins_final}
	\begin{split}
		\mathcal{L}_{t}(\bm{\Theta}_{t})- \mathcal{L}_{t}(\bm{\Theta}^\star) 
		&\leq \frac{\sigma_{\max}(\eta_t\mathbf{I} +  \frac{\rho}{2}\mathbf{S}_{+}\mathbf{S}_{+}^\top)}{2} \Big(\|\bm{\Theta}_{t} -\bm{\Theta}^\star\|_F^2 \\
		&- \|\bm{\Theta}_{t+1} - \bm{\Theta}^\star \|_F^2\Big) + (c_1 + 4 c_\mathcal{N})\| \bm{\beta}_{t}\|_F^2  \\
		&+ (c_2+ c_\mathcal{N}) \| \bm{\beta}_{t-1}\|_F^2 + \frac{c_\mathcal{N} }{\sigma^2_{\max}(\mathbf{S}_{-}) }\|\partial \mathcal{L}_{t} (\bm{\Theta}_{t})\|_F^2 \\
		&+ ( \frac{\eta_2}{2} + \frac{\eta_3}{2} +  \frac{\rho\eta_5}{4})\sigma^2_{\max}(\mathbf{S}_{-}) \|\bm{E}_{t} \|_F^2,
	\end{split}
\end{equation}
where $c_1, c_2$ and $c_\mathcal{N}$ are defined as follows:
\begin{equation}
	\begin{split}
		c_1 & :=  \frac{1}{2\eta_2} -\frac{4}{\rho} + \frac{\eta_4}{2} +  \frac{1}{\rho\eta_1} + \frac{1}{2\eta_3} + \frac{2}{\rho \eta_6} - \frac{1}{2\eta_2\eta_6} - \frac{\eta_4}{2\eta_6}, \\
		c_2& := \frac{\eta_1}{\rho}-\frac{2}{\rho} + \frac{1}{2\eta_2} + \frac{\eta_4}{2} + \frac{2\eta_6}{\rho } - \frac{\eta_6}{2\eta_2} - \frac{\eta_4\eta_6}{2},\\
		c_\mathcal{N}& := \Big(\frac{\sigma_{\max}(\eta_t\mathbf{I} +  \frac{\rho}{2}\mathbf{S}_{+}\mathbf{S}_{+}^\top)}{2}+  \frac{\sigma^2_{\max}(\mathbf{S}_{-})}{2\eta_4} + \frac{\rho\sigma^2_{\max}(\mathbf{S}_{-})}{4\eta_5} \Big) \\
		&  \qquad   \frac{\sigma^2_{\max}(\mathbf{S}_{-})}{\sigma^2_{\min}(\eta_t\mathbf{I} + 2\rho\bm{D})}.
	\end{split}\nonumber
\end{equation} 
Carefully choose $\eta_1, \eta_2, \eta_3, \eta_4, \eta_5$, and $\eta_6$, we can make $c_1 + 4c_\mathcal{N} = -(c_2+c_\mathcal{N}) = -c$, where $c>0$. Then \eqref{eq:stat_ins_final} can be further simplified as
\begin{equation}
	\label{eq:stat_ins_final_simp}
	\begin{split}
		&\mathcal{L}_{t}(\bm{\Theta}_{t})- \mathcal{L}_{t}(\bm{\Theta}^\star) \\
		&\leq \frac{\sigma_{\max}(\eta_t\mathbf{I} +  \frac{\rho}{2}\mathbf{S}_{+}\mathbf{S}_{+}^\top)}{2} \Big(\|\bm{\Theta}_{t} -\bm{\Theta}^\star\|_F^2 - \|\bm{\Theta}_{t+1} - \bm{\Theta}^\star \|_F^2\Big) \\
		&+  c (\|\bm{\beta}_{t-1}\|_F^2 -\| \bm{\beta}_{t}\|_F^2)   + \frac{c_\mathcal{N} }{\sigma^2_{\max}(\mathbf{S}_{-}) }\|\partial \mathcal{L}_{t} (\bm{\Theta}_{t})\|_F^2 \\ 
		&+ ( \frac{\eta_2}{2} + \frac{\eta_3}{2} +  \frac{\rho\eta_5}{4})\sigma^2_{\max}(\mathbf{S}_{-}) \|\bm{E}_{t} \|_F^2.
	\end{split}
\end{equation}

To satisfy $c_1 + 4c_\mathcal{N} = -(c_2+c_\mathcal{N})$, one example is to set $\eta_4 = \frac{2\eta_6}{(\eta_6-1)^2}( \frac{1}{\rho}(\eta_1 + \frac{1}{\eta_1} + \frac{2}{\eta_6} + 2\eta_6) + \frac{1}{2\eta_2}(2-\eta_6 -\frac{1}{\eta_6}) + \frac{1}{2\eta_3} + 5c_\mathcal{N})$.

Summarizing both sides of \eqref{eq:stat_ins_final_simp} from $t=1$ to $t=T$ leads to the accumulated network regret $\mathcal{R}(T)$:
\begin{equation}
	\label{eq:stat_reg_bound}
	\begin{split}
		&\quad \mathcal{R}(T) \\
		&\leq \frac{\sigma_{\max}(\eta_t\mathbf{I} +  \frac{\rho}{2}\mathbf{S}_{+}\mathbf{S}_{+}^\top)}{2} \Big( \|\bm{\Theta}_1-\bm{\Theta}^\star \|_F^2 - \| \bm{\Theta}_{T+1} -\bm{\Theta}^\star\|_F^2 \Big) \\
		& + c(\|\bm{\beta}_{0}\|_F^2 -\| \bm{\beta}_{T}\|_F^2 )  +  \sum_{t=1}^T \frac{c_\mathcal{N} }{\sigma^2_{\max}(\mathbf{S}_{-}) }\|\partial \mathcal{L}_{t} (\bm{\Theta}_{t})\|_F^2 \\
		&+ \sum_{t=1}^T ( \frac{\eta_2}{2} + \frac{\eta_3}{2} +  \frac{\rho\eta_5}{4})\sigma^2_{\max}(\mathbf{S}_{-}) \|\bm{E}_{t} \|_F^2\\
		&\leq \frac{\sigma_{\max}(\eta_t\mathbf{I} +  \frac{\rho}{2}\mathbf{S}_{+}\mathbf{S}_{+}^\top)}{2}  \|\bm{\Theta}_1-\bm{\Theta}^\star \|_F^2 + c\|\bm{\beta}_{0}\|_F^2\\
		& + \sum_{t=1}^T \frac{c_\mathcal{N} }{\sigma^2_{\max}(\mathbf{S}_{-}) }\|\partial \mathcal{L}_{t} (\bm{\Theta}_{t})\|_F^2 \\
		&+ \sum_{t=1}^T ( \frac{\eta_2}{2} + \frac{\eta_3}{2} +  \frac{\rho\eta_5}{4})\sigma^2_{\max}(\mathbf{S}_{-}) \|\bm{E}_{t} \|_F^2\\
		&=\frac{\sigma_{\max}(\eta_t\mathbf{I} +  \frac{\rho}{2}\mathbf{S}_{+}\mathbf{S}_{+}^\top)}{2}  \|\bm{\Theta}^\star \|_F^2  \\
		&+ \sum_{t=1}^T \frac{c_\mathcal{N} }{\sigma^2_{\max}(\mathbf{S}_{-}) }\|\partial \mathcal{L}_{t} (\bm{\Theta}_{t})\|_F^2 \\
		&+ \sum_{t=1}^T ( \frac{\eta_2}{2} + \frac{\eta_3}{2} +  \frac{\rho\eta_5}{4})\sigma^2_{\max}(\mathbf{S}_{-}) \|\bm{E}_{t} \|_F^2.
	\end{split}
\end{equation}
The last equality comes from the initialization that $\bm{\Theta}_{1}=\mathbf{0}$ and $\bm{\beta}_1=\mathbf{0}$ and thus $\bm{\beta}_0=\mathbf{0}$. Assumption \ref{ass:boundtheta} assumes $\|\bm{\theta}^\star\|_F\leq C_{\theta}$, which implies $\|\bm{\Theta}^\star\|_F\leq \sqrt{N}C_{\theta}$. Assumption \ref{ass:convex} assumes $\|\partial\mathcal{L}_{t}(\bm{\Theta}_{t})\|_F\leq  C_{\mathcal{L}}$. Then, setting $\rho  = \eta_t = \eta_2 =\eta_3 = 1/\mathcal{O}(\sqrt{T})$, the sublinear regret is achieved:
\begin{equation}
	\label{eq:stat_reg_bound_2}
	\mathcal{R}(T) \leq (\sqrt{N}C_{\theta} + \frac{1}{\sigma^2_{\max}(\mathbf{S}_{-})}  C_{\mathcal{L}} + \sigma^2_{\max}(\mathbf{S}_{-})\zeta)\mathcal{O}(\sqrt{T}),
\end{equation}
where  $\zeta:=\max\{\sqrt{N}\alpha\beta, \sqrt{2NL}\Delta/2\}$ with $\alpha$ and $\beta$ being the predefined censoring threshold parameters and $\Delta$ being the length of the quantization interval.

\section{References Section}
\bibliography{references}
\bibliographystyle{IEEE}

\newpage

% your own custom macro containing the $\backslash${\tt{includegraphics}} command to make things
%\bf{If you include a photo:}\vspace{-33pt}
%\begin{IEEEbiography}[{\includegraphics[width=1in,height=1.25in,clip,keepaspectratio]{fig1}}]{Michael Shell}
%Use $\backslash${\tt{begin\{IEEEbiography\}}} and then for the 1st argument use $\backslash${\tt{includegraphics}} to declare and link the author photo.
%Use the author name as the 3rd argument followed by the biography text.
%\end{IEEEbiography}

\vfill

\end{document}